%% file: main.tex
\documentclass[11pt]{article}

\usepackage[preprint]{acl}

\usepackage{times}
\usepackage{latexsym}

\usepackage[T1]{fontenc}

\usepackage[utf8]{inputenc}

\usepackage{microtype}

\usepackage{inconsolata}

\usepackage{graphicx}

\usepackage{booktabs}
\usepackage{multirow}
\usepackage{rotating}
\usepackage{xspace}
\usepackage{amsmath}
\usepackage{amssymb}
\usepackage{subcaption}
\usepackage{tikz}
\usepackage{xcolor}
\usetikzlibrary{arrows.meta,positioning,fit,backgrounds,decorations.pathreplacing,calc}
\newcommand{\FullMethod}{\textbf{C}ross-Lingual \textbf{A}lignment of \textbf{R}epresentations in a Language-Agnostic Space via \textbf{O}ptimal \textbf{T}ransport\xspace}
\newcommand{\OurMethod}{CAROT\xspace}
\newcommand{\mathvec}[1]{\mathbf{#1}}

\title{Cross-Lingual Representation Alignment by Token-Level Optimal Transport in a Language-Agnostic Space}

\author{
  \textbf{Taisei Yamamoto\textsuperscript{1,2}},
  \textbf{Ryoma Kumon\textsuperscript{1,2}},
  \textbf{Danushka Bollegala\textsuperscript{3}},
  \textbf{Hitomi Yanaka\textsuperscript{1,2,4}}\\
  \textsuperscript{1}The University of Tokyo 
  \textsuperscript{2}Riken 
  \textsuperscript{3}University of Liverpool
  \textsuperscript{4}Tohoku University\\
  \texttt{\{yamamo96, kumoryo9, hyanaka\}@is.s.u-tokyo.ac.jp}\\
  \texttt{danushka@liverpool.ac.uk}
}

\begin{document}
\maketitle
\begin{abstract}
Cross-lingual alignment (CLA) aims to align the representations of large language models (LLMs) across languages, enabling cross-lingual transfer to improve multilingual capabilities.
Previous CLA methods often ignore language-specific information encoded in representations and only consider sentence-level alignment, which may lead to suboptimal performance and input-output language mismatch.
We propose \textbf{\OurMethod} (\FullMethod), which consists of two steps: identifying language-specific representations in LLMs' internal states and aligning language-agnostic representations across languages at the token level by optimal transport, while explicitly preserving language-specific representations.
Inference-time steering experiments show that the representations computed by \OurMethod are effective alignment targets, improving multilingual performance by up to 11.2 points in accuracy while maintaining input-output language consistency.
We further use the representations obtained by \OurMethod as training targets, internalizing the aligned representations.
The trained models outperform existing CLA methods in 11 of 18 evaluation settings (3 models $\times$ 3 tasks $\times$ ID/OOD languages).
Our work provides insights into what constitutes effective alignment targets for CLA in LLMs.
Code is available at \url{https://github.com/ynklab/CAROT}.
\end{abstract}

\input{chapters/01_introduction}

\input{chapters/02_related_work}

\input{chapters/03_method}

\input{chapters/setup.tex}

\input{chapters/04_eval_analysis}

\input{chapters/05_training}

\input{chapters/06_conclusion}

\section*{Limitations}

Our experiments are conducted on 4--8B parameter models due to computational resource constraints.
As \citet{lim2025languagespecificlatentprocesshinders} suggest that the multilingual mechanisms of LLMs may differ by model size, the effectiveness of our method on larger models remains to be investigated.

In \OurMethod, we use LEACE to identify language-specific and language-agnostic components in the model's representations.
LEACE considers only linear features, while LLMs involve non-linear transformations.
Thus, it is possible that \OurMethod may overlook non-linear language features that LLMs actually use, which may limit the effectiveness of our method.

As discussed in \S\ref{subsec:training_results}, our training relies on SFT-based objectives, while recent training approaches usually adopt RL to improve reasoning capabilities.
We expect that \OurMethod can be incorporated into RL frameworks by alternately applying $\mathcal{L}_\text{align}$ and the RL loss at each training step, which is left for future work.

Additionally, as discussed in \S\ref{subsec:training_results}, we use the precomputed LEACE and covariance matrices obtained from the base model throughout training.
However, the model's representations drift during training, as shown by the large principal angles between the pre- and post-training LEACE and covariance matrices.
Although we confirm that language-agnostic representations are more aligned between the source and target languages after training (\S\ref{subsec:training_results}), periodically updating the LEACE and covariance matrices during training may provide a more accurate training signal, which is also left for future work.

\citet{han2025rethinkingcrosslingualalignmentbalancing} point out that cross-lingual alignment often sacrifices the model's ability to answer culturally specific questions, as representations are aligned to the dominant language.
Meanwhile, \OurMethod preserves language-specific representations, which may help mitigate this issue.
We do not evaluate the model's cultural understanding in this work, and leave it for future work.

\section*{Acknowledgments}

We would like to thank the three anonymous reviewers and the meta-reviewer for their thoughtful engagement with our work and constructive feedback.
We also thank Adam Nohejl, Xiaotian Wang, and Koki Ryu for their valuable discussions and comments on the paper.
This work was supported by JST CREST Grant Number JPMJCR2565, Japan, and JST BOOST Program Grant Number JPMJBY24H5, Japan.

\bibliography{custom}

\appendix

\input{chapters/appendix.tex}

\end{document}

%% file: chapters/01_introduction.tex
\section{Introduction}
\label{sec:introduction}

\input{figures/sources/intro/figure_overview.tex}

As large language models (LLMs) are being used around the world, it is crucial that LLMs possess multilingual capabilities.
However, previous studies have shown that LLMs have inconsistent abilities across languages, especially showing limited understanding of low-resource languages~\citep[e.g.,][]{ahuja-etal-2023-mega,singh-etal-2025-global}.
As low-resource languages lack sufficient training data, it is important to achieve cross-lingual transfer, the ability of LLMs to generalize the knowledge and skills learned in one language to others.
One key factor enabling cross-lingual transfer is cross-lingual alignment (CLA).
CLA aligns the representations of LLMs across languages, allowing the model to map semantically similar inputs from different languages onto similar representations.

Previous CLA methods are mainly based on insights into the multilingual mechanisms of LLMs.
Several studies have suggested that LLMs follow a three-stage pipeline when processing multilingual inputs: (1) mapping inputs into a semantic space, (2) performing inference in that semantic space, and (3) translating back into the output language for generation~\citep{wu2025the,zhao2024how,tang-etal-2024-language}.
Multiple studies have reported a correlation between task performance in a given language and the similarity of its internal representations to those of the dominant language~\citep{kargaran-etal-2025-mexa,Li_Shi_Liu_Yang_Payani_Liu_Du_2025}.
These findings have motivated the development of CLA methods aimed at increasing the cross-lingual similarity of the middle-layer representations of LLMs through inference-time steering~\citep{zhao2026when,li-etal-2026-unlocking} and training~\citep{liu-niehues-2025-middle,li-etal-2024-improving-context,zhao2025lensrethinkingmultilingualenhancement}.

To promote CLA by leveraging representations in high-resource languages and identifying ideal representations for multilingual inputs, we must address two problems: \emph{what} to align and \emph{where} to align it.
Most training methods pull the hidden states of parallel sentences across languages closer together, but they do not necessarily consider the language-specific information encoded in the representations~\citep{liu-niehues-2025-middle,li-etal-2024-improving-context}.
Indiscriminately minimizing cross-lingual distance risks destroying the language-specific features models rely on, so language-specific components should be protected while aligning the language-agnostic components.
Furthermore, many steering methods bring the multilingual representations closer to the model's central language by applying a global vector shift, but they do not perform token-level alignment~\citep{zhao2026when,li-etal-2026-unlocking}.
As languages differ in morphological structure and grammatical constructions, a finer-grained token-level investigation of multilingual representations is necessary to develop more effective CLA methods.
For example, \textit{``I like dogs''} in English and \textit{``Watashi wa inu ga suki desu''} in Japanese (\autoref{fig:calot_overview}) have different word orders, and the Japanese sentence contains a topic marker ``wa'' and a subject marker ``ga'' that are not present in the English sentence.

To overcome these challenges, we propose \textbf{\OurMethod} (\FullMethod), a method to compute hidden representations that enhance the multilingual capabilities of LLMs by aligning token-level language-agnostic representations while preserving language-specific representations (\autoref{fig:calot_overview}).
We utilize LEACE~\citep{belrose2023leace} to find linearly encoded language-specific information (\S\ref{subsec:lang_decomp}) and isolate the language-agnostic representations.
We then align the language-agnostic representations between the source and target languages at the token level via optimal transport~\citep{chizat2018scaling,arase-etal-2023-unbalanced,dou-neubig-2021-word}, with dedicated adjustment for applying to LLMs' internal representations (\S\ref{subsec:token_align}).
The source language is the dominant language in which the model performs best (e.g., English), and the target language is the language we want to improve.
We confirm that the components removed by LEACE capture language-specific information and the remaining components correspond to language-agnostic semantic representations (\S\ref{sec:eval_lang_repr}).
Our steering experiments, covering a broad range of tasks, demonstrate that the representations obtained by \OurMethod are effective alignment targets, significantly improving multilingual performance without inducing input-output language mismatch (\S\ref{sec:steering_results}).

Inference-time steering with \OurMethod, however, requires a source-language translation of the input, which makes it impractical to apply at scale.
We therefore propose a training method that encourages the model's internal representations to match those obtained by \OurMethod (\S\ref{sec:training}), so that translation is required only during training.
Models trained with \OurMethod outperform existing CLA methods in 11 of 18 model-task-language settings (3 models $\times$ 3 tasks $\times$ ID/OOD languages), although the improvements do not yet match those observed with steering.
Our work provides insights into the ideal multilingual representations and an interpretable framework for achieving CLA in LLMs, contributing to more effective cross-lingual transfer.

%% file: figures/sources/intro/figure_overview.tex
\begin{figure*}[t]
    \centering
    \includegraphics[width=0.9\textwidth]{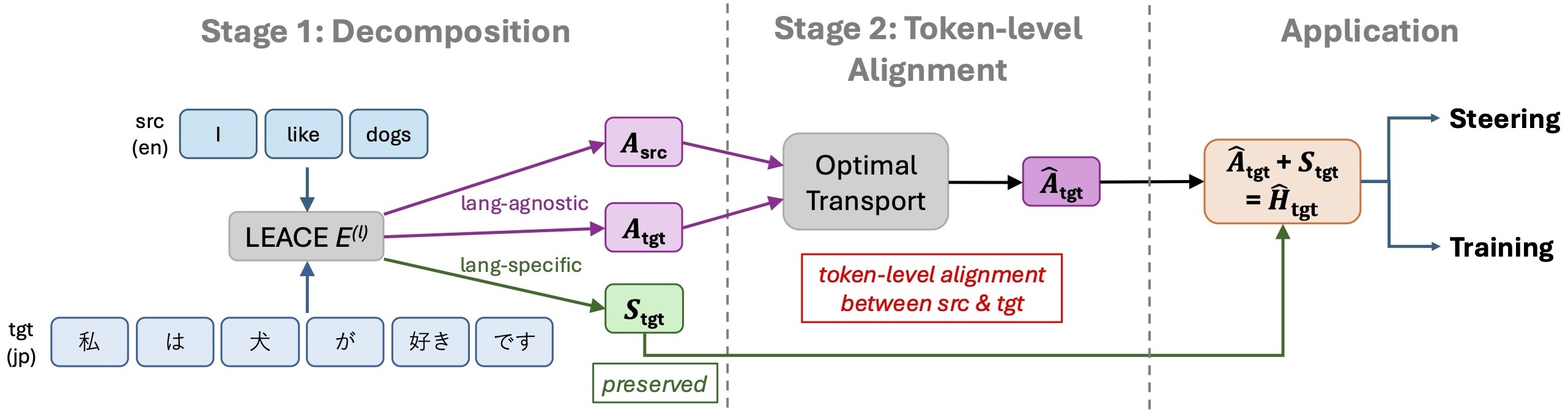}
    \caption{Overview of \OurMethod.
\textbf{Stage 1 (Decomposition):} LEACE decomposes the hidden states of each token into language-agnostic components ($\mathbf{A}_{\mathrm{src}}$ and $\mathbf{A}_{\mathrm{tgt}}$) and a language-specific component ($\mathbf{S}_{\mathrm{tgt}}$).
\textbf{Stage 2 (Alignment):} Unbalanced optimal transport aligns $\mathbf{A}_{\mathrm{src}}$ with $\mathbf{A}_{\mathrm{tgt}}$ at the token level, producing ideal language-agnostic representations $\hat{\mathbf{A}}_{\mathrm{tgt}}$.
\textbf{Application:} The aligned representations are recombined with the preserved language-specific component to form $\hat{\mathvec{H}}_{\mathrm{tgt}}$, which is used for inference-time steering or as a training signal.}
\label{fig:calot_overview}
\end{figure*}

%% file: chapters/02_related_work.tex
\section{Related Work}
\label{sec:related_work}

\paragraph{Multilingual Mechanisms of LLMs.}

A number of studies have investigated the multilingual mechanisms of LLMs, revealing that LLMs process multilingual inputs through a shared semantic space in middle layers.
\citet{wendler-etal-2024-llamas} and \citet{wu2025the} applied logit lens to the middle layers of LLMs and found that the top-ranked language is often the dominant language of the model.
\citet{chang-etal-2022-geometry} demonstrated that languages share common linear subspaces in the middle layers of multilingual LLMs.
These findings support the hypothesis that LLMs first map multilingual inputs into a shared semantic space and then perform inference in that space.

\paragraph{Cross-Lingual Alignment (CLA).}

In addition to the multilingual mechanisms of LLMs described above, \citet{kargaran-etal-2025-mexa} and \citet{Li_Shi_Liu_Yang_Payani_Liu_Du_2025} demonstrated that LLMs perform better in a language for which the hidden representations are closer to those of the dominant language.
Based on these findings, several studies applied inference-time steering to bring the multilingual representations closer to those of the dominant language~\citep{zhao2026when,lim2025languagespecificlatentprocesshinders}.
Furthermore, \citet{liu-niehues-2025-middle} and \citet{li-etal-2024-improving-context} proposed contrastive training methods to increase the similarity of the middle-layer representations of LLMs across languages.
However, these methods only consider sentence-level alignment, and most training methods ignore language-specific information encoded in representations.
Although \citet{zhao2025lensrethinkingmultilingualenhancement} proposed a cross-lingual training method that took into account language-specific representations, they only calculated and aligned sentence-level representations.
Token-level cross-lingual alignment has also been studied for word alignment, with several methods using optimal transport~\citep{dou-neubig-2021-word,arase-etal-2023-unbalanced}, but not for CLA in LLMs.
Our approach addresses both issues by aligning token-level language-agnostic representations.

\paragraph{Concept Erasure.}

Concept erasure identifies and removes specific information from representations without compromising other features~\citep{ravfogel-etal-2020-null,pmlr-v162-ravfogel22a}.
\citet{belrose2023leace} developed LEACE, a concept erasure method that computes an affine transformation that removes the linear features representing the target concept (\autoref{app:leace}).
They provided theoretical guarantees that after applying LEACE, the target labels cannot be classified linearly while minimizing the distance between the original and transformed representations.
In the multilingual setting, earlier work removed language-specific components from multilingual encoders in a predefined form, by subtracting per-language centroids~\citep{libovicky-etal-2020-language}, normalizing language-specific means and variances~\citep{zhao-etal-2021-inducing}, or projecting out low-rank language subspaces~\citep{xie-etal-2022-discovering,zhao2026when}.
We utilize LEACE to isolate language-specific features from the internal representations of LLMs.

%% file: chapters/03_method.tex
\section{\OurMethod}
\label{sec:method}

We propose \OurMethod, a method for computing multilingual representations of LLMs to enhance CLA.
\OurMethod consists of two stages: (1) decomposing hidden representations into language-specific and language-agnostic components using LEACE (\S\ref{subsec:lang_decomp}), and (2) aligning the language-agnostic representations of the source and target languages at the token level via optimal transport (\S\ref{subsec:token_align}).

\subsection{Language-Specific and Language-Agnostic Decomposition}
\label{subsec:lang_decomp}

\paragraph{Sentence Representations.}
To identify features encoding language information in LLMs' internal representations, we calculate sentence-level embeddings and train LEACE on these representations.
Specifically, we calculate the sentence representation $\mathvec{x}^{(l)}$ (at layer $l$) by aggregating the hidden states $\mathvec{h}_t^{(l)}$ using a weighted average across token positions $t$, following \citet{muennighoff2022sgptgptsentenceembeddings}:
\begin{equation}
    \label{eq:sentence_repr}
    \mathvec{x}^{(l)} = \sum_{t=1}^{n} w_t \mathvec{h}_t^{(l)}, \quad \text{where} \quad w_t = \frac{t}{\sum_{i=1}^{n} i}
\end{equation}
and $n$ is the number of tokens in the sentence.
This aggregation averages out token-level variation to expose the dominant semantic content across parallel sentence pairs, leaving the language direction more salient for LEACE to identify.

\paragraph{LEACE-Based Erasure.}
As described in \S\ref{sec:related_work}, LEACE calculates an affine transformation after which the target labels cannot be linearly separated.
We train LEACE in a multiclass setting, where the target labels are the languages of the input sentences.
LEACE is trained at each layer $l$, denoted as a function $E^{(l)}$.
LEACE is subsequently applied to the hidden vectors of each token.

\subsection{Token-Level Alignment with Optimal Transport in a Language-Agnostic Space}
\label{subsec:token_align}

\OurMethod aligns the token-level language-agnostic representations of the target language with those of the source language, while preserving the language-specific representations of the target language.
We select English as the source language as it is the dominant language for all models in our experiments and thus serves as a good anchor for alignment.

\paragraph{LEACE-Based Decomposition.}
Let $\mathvec{H}_\text{src} \in \mathbb{R}^{n_\text{src} \times d}$ and $\mathvec{H}_\text{tgt} \in \mathbb{R}^{n_\text{tgt} \times d}$ be the hidden states of the parallel sentences at layer $l$, where $n_\text{src}$ and $n_\text{tgt}$ are the number of tokens in the source and target language sentences, respectively, and $d$ is the hidden dimension.
We first apply the LEACE eraser $E^{(l)}$ to separate language-specific and language-agnostic representations:
\begin{align}
    \mathvec{A}_\text{src} &= E^{(l)}(\mathvec{H}_\text{src}), \;
    \mathvec{A}_\text{tgt} = E^{(l)}(\mathvec{H}_\text{tgt}), \\
    \mathvec{S}_\text{tgt} &= \mathvec{H}_\text{tgt} - \mathvec{A}_\text{tgt},
\end{align}
where $E^{(l)}$ is applied to each row of $\mathvec{H}$, i.e., each token's hidden vector.
$\mathvec{A}$ denotes the language-agnostic representation, and $\mathvec{S}_\text{tgt}$ is the language-specific component of the target language.

\paragraph{Optimal Transport in Language-Agnostic Space.}
We align source and target language-agnostic representations via optimal transport (OT), using Mahalanobis distance as the transport cost.
We precompute the covariance matrix $\mathbf{\Sigma}$ of sentence representations (\autoref{eq:sentence_repr}) on the same training data as LEACE.
We define the cost matrix $\mathvec{C} \in \mathbb{R}^{n_\text{tgt} \times n_\text{src}}$ as the squared Mahalanobis distance:
\begin{equation}
    C_{ij} = \bigl\|\mathbf{\Sigma}^{-1/2}(\mathvec{a}_{\text{tgt},i} - \mathvec{a}_{\text{src},j})\bigr\|_2^2,
\end{equation}
where $\mathvec{a}_{\text{tgt},i}^\top$ and $\mathvec{a}_{\text{src},j}^\top$ are the $i$-th and $j$-th row of $\mathvec{A}_\text{tgt}$ and $\mathvec{A}_\text{src}$, which correspond to the language-agnostic representations of the $i$-th target and $j$-th source token, respectively.
We use Mahalanobis distance considering that the hidden vectors of LLMs tend to be anisotropic\footnote{We confirm that especially gemma3-4b has highly anisotropic distributions of hidden representations, as discussed in \autoref{app:additional_lang_repr_results}.}~\citep{machina-mercer-2024-anisotropy}.

Since the source and target sequences generally differ in length ($n_\text{src} \neq n_\text{tgt}$) and not every token has a clear cross-lingual counterpart, we solve the \emph{unbalanced} OT problem with the Sinkhorn algorithm\footnote{\url{https://github.com/PythonOT/POT}}~\citep{chizat2018scaling,flamary2021pot} under uniform marginal distributions $\mathvec{p} = \tfrac{1}{n_\text{tgt}}\mathvec{1}$ and $\mathvec{q} = \tfrac{1}{n_\text{src}}\mathvec{1}$:
\begin{equation}
    \label{eq:sinkhorn}
    \mathvec{T} = \operatorname{Sinkhorn\text{-}UB}(\mathvec{p},\, \mathvec{q},\, \mathvec{C};\; \epsilon_{\mathrm{OT}},\, \rho),
\end{equation}
where $\epsilon_{\mathrm{OT}}$ controls entropic regularization and $\rho$ relaxes the marginal constraints.

\paragraph{Perplexity-Based Filtering.}
The transport plan $\mathvec{T}$ may assign diffuse mass to tokens that are difficult to align.
To suppress such noise, we compute the perplexity of the row-normalized transport distribution for each target token $i$:
\begin{align}
    \label{eq:perplexity}
    \mathrm{ppl}_i^{\mathrm{row}} = \exp\!\Bigl(-\sum_j \hat{T}_{ij}\log \hat{T}_{ij}\Bigr), \\
    \text{where} \quad \hat{T}_{ij} = \frac{T_{ij}}{\sum_j T_{ij} + \epsilon_{\mathrm{num}}}, \notag
\end{align}
and analogously for columns (source tokens).
$\epsilon_{\mathrm{num}}$ is a small constant for numerical stability.
Rows with $\mathrm{ppl}_i^{\mathrm{row}} > \theta \cdot n_\text{src}$ and columns with $\mathrm{ppl}_j^{\mathrm{col}} > \theta \cdot n_\text{tgt}$ are zeroed out, where $\theta$ is a hyperparameter.
This filtering retains only those target tokens that are confidently matched to a focused subset of source tokens, and vice versa.

\paragraph{Transported Representation.}
For each target token $i$, the received transport mass $m_i = \sum_j T_{ij}$ quantifies alignment confidence.
We compute the new language-agnostic representation as a soft interpolation between the transported source and the original target:
\begin{equation}
    \hat{\mathvec{A}}_\text{tgt} = \boldsymbol{\alpha} \odot \hat{\mathvec{T}}\mathvec{A}_\text{src} + (\mathvec{1} - \boldsymbol{\alpha}) \odot \mathvec{A}_\text{tgt},
\end{equation}
where $\hat{\mathvec{T}}$ is the row-normalized filtered transport plan and $\alpha_i = \min(1, m_i/p_i)$ scales each target token's update proportionally to its received mass.
Tokens with little or no received mass (e.g., by being filtered out) thus retain their original language-agnostic representation.

\paragraph{Language-Specific Component Restoration.}
Finally, we add back the preserved target language-specific component:
\begin{equation}
    \hat{\mathvec{H}}_\text{tgt} = \hat{\mathvec{A}}_\text{tgt} + \mathvec{S}_\text{tgt}.
\end{equation}
The resulting $\hat{\mathvec{H}}_\text{tgt}$ carries language-agnostic representations aligned with the source language while retaining the target language identity.

\subsection{Using \OurMethod Representations}
\label{subsec:using_repr}

We use $\hat{\mathvec{A}}_\text{tgt}$ and $\hat{\mathvec{H}}_\text{tgt}$ in two ways: as (1) an inference-time steering method and (2) a training target to encourage the model to produce better-aligned representations.

\paragraph{Inference-Time Application.}
When used for inference-time steering, \OurMethod is applied to the hidden states of the tokens corresponding to a question (and answer options for multiple-choice tasks) at a single layer, swapping in the new representations $\hat{\mathvec{H}}_\text{tgt}$ before the model continues forward.
BOS, EOS, and padding tokens are excluded from steering to preserve the model's generation behavior.
Note that steering requires a source-language translation of the input.
Our steering evaluation (\S\ref{sec:steering_results}) therefore serves to demonstrate the effectiveness of the representations computed by \OurMethod as alignment targets, while the approach itself remains applicable to offline settings where parallel inputs are naturally available, such as generating higher-quality reasoning traces in low-resource languages as distillation data.

\paragraph{Training Objective.}
When used for training, we fine-tune the model with two loss components: language-agnostic alignment loss $\mathcal{L}_\text{align}$ and language modeling loss $\mathcal{L}_\text{LM}$.
\emph{Language-agnostic alignment loss} is calculated on parallel data.
We penalize the deviation of the model's current language-agnostic component $\mathvec{A}_\text{tgt}$ from the \OurMethod-derived $\hat{\mathvec{A}}_\text{tgt}$ using the same Mahalanobis distance as OT cost:
\begin{equation}
    \mathcal{L}_\text{align} = \frac{1}{n_\text{tgt}} \sum_{i=1}^{n_\text{tgt}}
    \frac{\bigl\|\mathbf{\Sigma}^{-1/2}\,(\mathvec{a}_{\text{tgt},i} - \hat{\mathvec{a}}_{\text{tgt},i})\bigr\|_2^2}
         {\bigl\|\mathbf{\Sigma}^{-1/2}\,\hat{\mathvec{a}}_{\text{tgt},i}\bigr\|_2^2 + \epsilon_{\mathrm{num}}}.
\end{equation}
Normalizing the loss by the target norm stabilizes the magnitude of the loss.
\emph{Language modeling loss} is calculated on the instruction-tuning data.
We minimize the supervised fine-tuning (SFT) loss $\mathcal{L}_\text{LM}$ on the answer tokens of the target batch.
In the first half of training, we optimize the two losses alternately at each training step rather than combining them into a single loss to avoid interference between the two objectives.
In the second half, we only optimize $\mathcal{L}_\text{LM}$ to allow the model to leverage its aligned representations for generation.

Unlike steering, this application requires parallel data only during training.
As parallel corpora are available for many languages and our training method requires only a modest amount of parallel data, training with \OurMethod is a practical route to improving the multilingual ability of LLMs.

%% file: chapters/setup.tex
\section{Experimental Setup}
\label{sec:eval_settings}

\paragraph{Models and Datasets.}

In our experiments, we use three multilingual LLMs: \textit{Llama3.1-8B}~\citep{grattafiori2024llama3herdmodels}, \textit{gemma3-4b}~\citep{gemma_2025}, and \textit{Qwen3-4B}~\citep{qwen3technicalreport}.
For training LEACE, we use parallel sentences from two multilingual datasets: WMT24++~\citep{deutsch-etal-2025-wmt24} and NTREX~\citep{federmann-etal-2022-ntrex}.
We include 12 in-distribution (ID) languages for LEACE, which vary in terms of their linguistic characteristics and resource levels (see \autoref{tab:languages}).
In total, we use 2,560 sentences per language for training LEACE.

For evaluation, we use KLAR Continent category~\citep{wang-etal-2025-lost-multilinguality} as the knowledge retrieval task, GMMLU~\citep{singh-etal-2025-global} \textit{test} set as the task that requires both knowledge and reasoning, BELEBELE~\citep{bandarkar-etal-2024-belebele} for reading comprehension, and XQuAD~\citep{artetxe-etal-2020-cross} for open-ended question answering.
We also use FLORES+~\citep{flores_plus_nllb-24} for analyzing the internal representations of the models in \S\ref{sec:eval_lang_repr}.
The source language is English, and the target languages are seven languages for KLAR, nine for GMMLU and BELEBELE, and five for XQuAD, selected based on the availability of the datasets and the variety of characteristics (\autoref{tab:languages}).

\paragraph{Hyperparameter Search.}

In the steering and training experiments in \S\ref{sec:steering_results} and \S\ref{sec:training}, we perform a grid search to find the optimal hyperparameters for each method and model.
We use the \textit{dev} set of GMMLU and adopt the parameters achieving the best accuracy.
For steering with \OurMethod, we search for the best layer from every three layers, and the filtering threshold $\theta$ from $\{0.25, 0.5, 0.75, 1.0\}$ (note that $\theta$=1 means that no filter is applied).
For training with \OurMethod, we adopt the best layer and filtering threshold from the steering experiments for each model, and search for the best learning rate from \{5e-4, 1e-4, 5e-5, 1e-5\}.

\paragraph{Metrics.}

We use accuracy, (language) fidelity, and cross-lingual agreement ratio as evaluation metrics.
Fidelity is the percentage of instances where the model's answer is in the same language as the input.
We use GlotLID~\citep{kargaran-etal-2023-glotlid} to identify the language of the model's output.
Cross-lingual agreement ratio is the percentage of instances for which the model's answer in the target language matches that in the source language.
The details of the evaluation protocols are described in \autoref{app:eval_protocols}.

%% file: chapters/04_eval_analysis.tex
\section{Analysis of Language-Specific and Language-Agnostic Representations}
\label{sec:eval_lang_repr}

\input{figures/sources/leace/figure_llama_leace_scatter.tex}

\input{figures/sources/leace/figure_llama_space_cossim.tex}

First, we analyze the representations obtained by applying LEACE on sentence-level embeddings (\S\ref{subsec:lang_decomp}).
\autoref{fig:llama_leace_scatter_layer012} shows PCA-based and t-SNE-based~\citep{JMLR:v9:vandermaaten08a} 2D scatter plots of the original, remaining, and LEACE-erased sentence representations of Llama3.1-8B at layer 12, calculated on the FLORES+ \textit{devtest} set.
We show the results for 12 ID languages and 6 out-of-distribution (OOD) languages, which are not used for training LEACE.
For ID languages, sentences in the remaining representations are mixed across languages, while the LEACE-erased representations are clearly separated by language.
These observations confirm that LEACE successfully identifies language-specific representations.
For OOD languages, the remaining representations are somewhat clustered by language, indicating residual language information not captured by LEACE.
Meanwhile, the erased representations still show clear separation.
This suggests that LEACE captures language-specific representations that generalize to OOD languages to some extent.
The same trend holds across other layers and models (\autoref{app:additional_lang_repr_results}).

\autoref{fig:llama_leace_embedding_similarity} shows the average cosine similarity of the sentence representations between the intra-language pairs (i.e., sentences in the same language) and inter-language pairs (i.e., parallel sentence pairs between the source and target languages) for Llama3.1-8B, calculated on the FLORES+ \textit{devtest} set.
In the remaining representations, inter-language similarity is higher than intra-language similarity, showing that content rather than language determines the representation (i.e., the representations are language-agnostic).
In contrast, in the LEACE-erased representations, intra-language similarity is consistently close to 1, indicating that erased representations follow a common direction within the same language regardless of content (i.e., the representations are language-specific).
These results confirm that LEACE successfully disentangles language-specific and language-agnostic components.

\section{Steering Evaluation}
\label{sec:steering_results}

\subsection{Baselines}
We implement two steering methods for comparison with \OurMethod.
One is a vector offset based on diff-in-means~\citep{lim2025languagespecificlatentprocesshinders}, which we refer to as \textit{Lang Vec}.
The other is \textit{MLRS}~\citep{zhao2026when}, which first identifies language-specific components and then suppresses them in the middle layers and adds them back in the later layers.
Note that these methods perform sentence-level alignment, while \OurMethod is a token-level alignment method.
Also note that although MLRS considers language-specific representations, it does not give theoretical grounds for the decomposition, while \OurMethod does by using LEACE.
We describe the details of these baselines in \autoref{app:baseline_steering}.

\subsection{Results and Analysis}

\input{figures/sources/steering/table_steering_results.tex}

\autoref{tab:steering_aggregate} shows the task accuracy and language fidelity for each method, averaged across target languages.
\OurMethod outperforms the baselines in most cases, achieving a significant improvement of up to 11.2 points in accuracy over the original model (Unsteered).
Meanwhile, fidelity does not degrade substantially, indicating that \OurMethod does not cause input-output language mismatch.
For gemma3-4b on GMMLU, fidelity is lower than in the other settings for all methods, which may be due to the model's tendency to generate reasoning steps in English even when the input is in another language.
Additionally, \autoref{tab:agreement} in \autoref{app:additional_steering_results} reports the agreement ratio, showing that \OurMethod achieves the highest agreement in most cases.
These results indicate that \OurMethod improves CLA.

The best steering layers of \OurMethod are all in the early-to-middle layers (layer 12, 6, 9 for Llama3.1-8B, gemma3-4b, and Qwen3-4B, respectively, as shown in \autoref{app:selected_parameter}).
As the intervention targets only the question tokens, steering applied at later layers may have a weaker effect once the relevant information has already been propagated to the answer-generating position.

\input{figures/sources/steering/figure_llama_steering_results.tex}

\autoref{fig:llama_steering_results_gmmlu} shows layer-wise per-language fidelity and accuracy for Llama3.1-8B on the GMMLU \textit{dev} set when steering with \OurMethod, calculated when conducting hyperparameter search.
\OurMethod achieves high fidelity across all languages and layers, and the accuracy improves most in the early layers.
In contrast, the results for Lang Vec show degraded fidelity in many cases (\autoref{fig:llama_gemma_qwen_steering_results_lang_vec_gmmlu_dev}).
Lang Vec brings the hidden states of the target language closer to those of the source language, which can cause a loss of language-specific information and thus low fidelity.
Additional results are shown in \autoref{app:additional_steering_results}.

\input{figures/sources/steering/figure_llama_semantic_cossim.tex}

\autoref{fig:llama_semantic_cossim_alignment_steering} shows the cosine similarity of language-agnostic sentence representations between the source and target languages when steering with \OurMethod at layer 12 of Llama3.1-8B, calculated on the FLORES+ \textit{devtest} set.
Similarity increases at the steering layer, and higher similarity lasts in later layers.
This confirms that \OurMethod successfully aligns the language-agnostic representations between the source and target languages.
As a sanity check, we inspect several samples and confirm that transport plans are sparse and exhibit little noise, and that semantically corresponding tokens are correctly aligned across languages (\autoref{app:ot_plans}).

An ablation study (\autoref{tab:steering_ablation}, \autoref{app:additional_steering_results}) shows that LEACE-based decomposition and filtering of the OT plan contribute to the robustness of \OurMethod.
Removing LEACE-based decomposition degrades performance in most cases.
Removing the filtering yields comparable accuracy but is less stable (e.g., a drop of 14.0 points on Qwen3-4B/KLAR) and lowers fidelity in many settings.
In addition, we compare \OurMethod with a variant that swaps the language-agnostic representations of only the last token between the source and target languages to investigate the effect of token-level alignment.
The results show that \OurMethod outperforms this variant in most cases, indicating that token-level alignment effectively improves accuracy.

\begin{table}
    \centering
    \small
    \begin{tabular}{lccc}
        \toprule
        Lang & orig fwd & src fwd & \OurMethod fwd \\
        \midrule
        Spanish & 28.1 $\pm$ 2.6 & 24.3 $\pm$ 0.8 & 29.6 $\pm$ 4.1 \\
        Japanese & 28.7 $\pm$ 2.7 & 28.2 $\pm$ 2.9 & 30.8 $\pm$ 6.3 \\
        Telugu & 36.8 $\pm$ 15.8 & 24.4 $\pm$ 1.8 & 49.6 $\pm$ 30.0 \\
        \bottomrule
    \end{tabular}
    \caption{Computational overhead of \OurMethod measured on Llama3.1-8B with 30 samples from the GMMLU \textit{test} set.
We show the average time (in milliseconds) for one forward pass of the original model in the target language (\textit{orig fwd}), the original model in the source language (\textit{src fwd}), and the model steered with \OurMethod (\textit{\OurMethod fwd}).
The average number of tokens in the source language (English) is 83.1, and in the target languages, 122.2 (Spanish), 122.1 (Japanese), and 601.4 (Telugu).
}
    \label{tab:steering_overhead}
\end{table}

The computational complexity of \OurMethod is $O(n_{\text{src}} n_{\text{tgt}} (d + T))$, where $T$ is the number of iterations for Sinkhorn's algorithm (we set $T=1000$).
\autoref{tab:steering_overhead} shows the average per-forward-pass latency for Llama3.1-8B on the GMMLU \textit{test} set (30 samples per language).
\OurMethod adds a second, source-language forward pass and the OT computation, roughly doubling the prefill cost (e.g., $24.3+29.6=53.9$ ms vs.\ $28.1$ ms for Spanish).
However, since \OurMethod is applied only once at prefill, this cost is amortized over the entire generation process, which involves as many forward passes as the number of tokens generated.
The overhead of \OurMethod is therefore modest in practice, especially for long generations.

%% file: figures/sources/leace/figure_llama_leace_scatter.tex
\begin{figure}[t]
  \centering
  \begin{minipage}{0.95\columnwidth}
    \centering
    \includegraphics[width=\columnwidth]{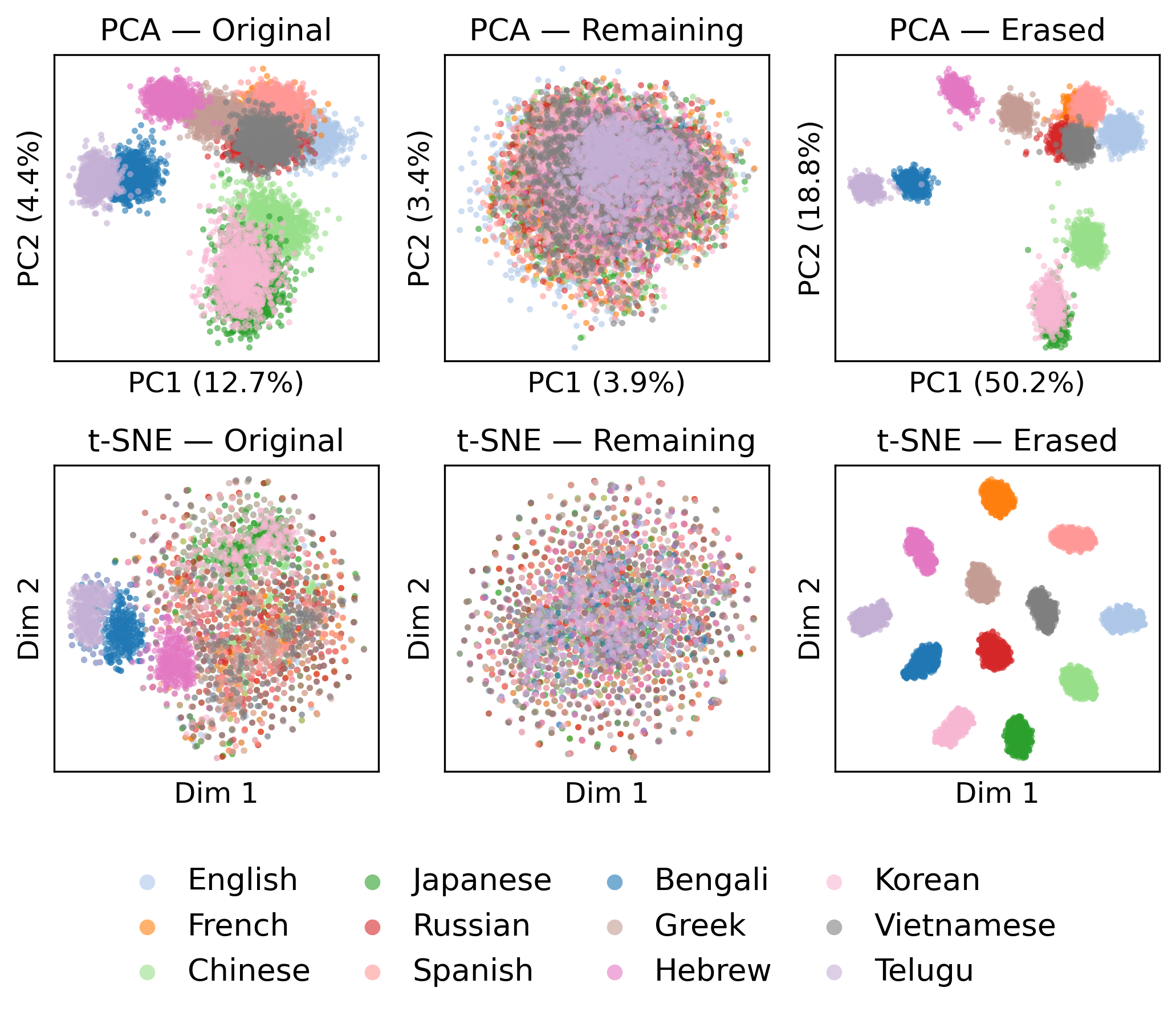}
    \subcaption{In-distribution languages}
    \label{fig:llama_leace_scatter_layer012_id}
  \end{minipage}
  \hspace{-3mm}
  \begin{minipage}{0.95\columnwidth}
    \includegraphics[width=\columnwidth]{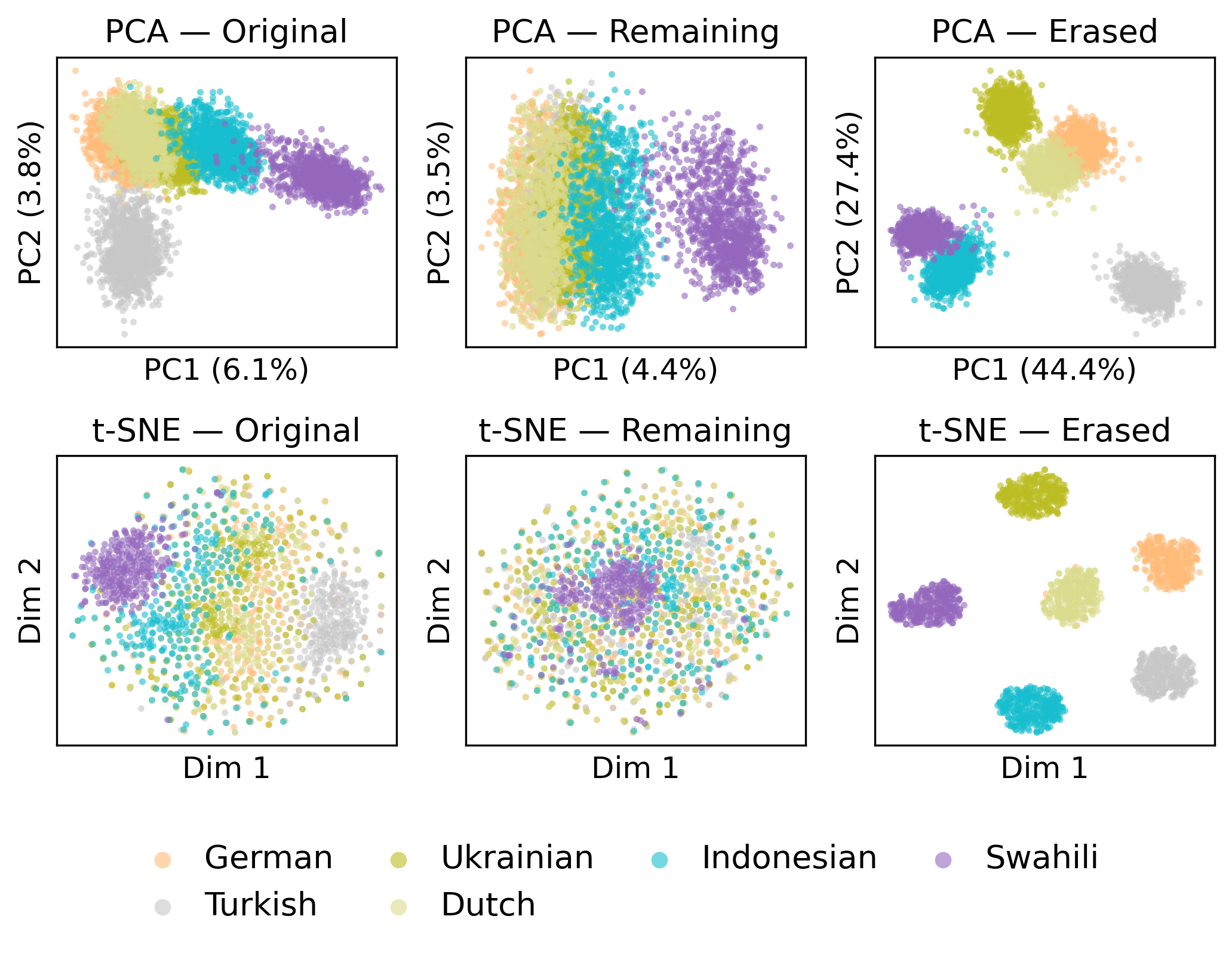}
    \subcaption{Out-of-distribution languages}
    \label{fig:llama_leace_scatter_layer012_ood}
  \end{minipage}
  \caption{Scatter plot of original (left), remaining (center), and LEACE-erased (right) representations of Llama3.1-8B at layer 12.
  Each point represents the sentence embedding, colored by language.}
  \label{fig:llama_leace_scatter_layer012}
\end{figure}

%% file: figures/sources/leace/figure_llama_space_cossim.tex
\begin{figure*}[t]
    \centering
    \includegraphics[width=.85\linewidth]{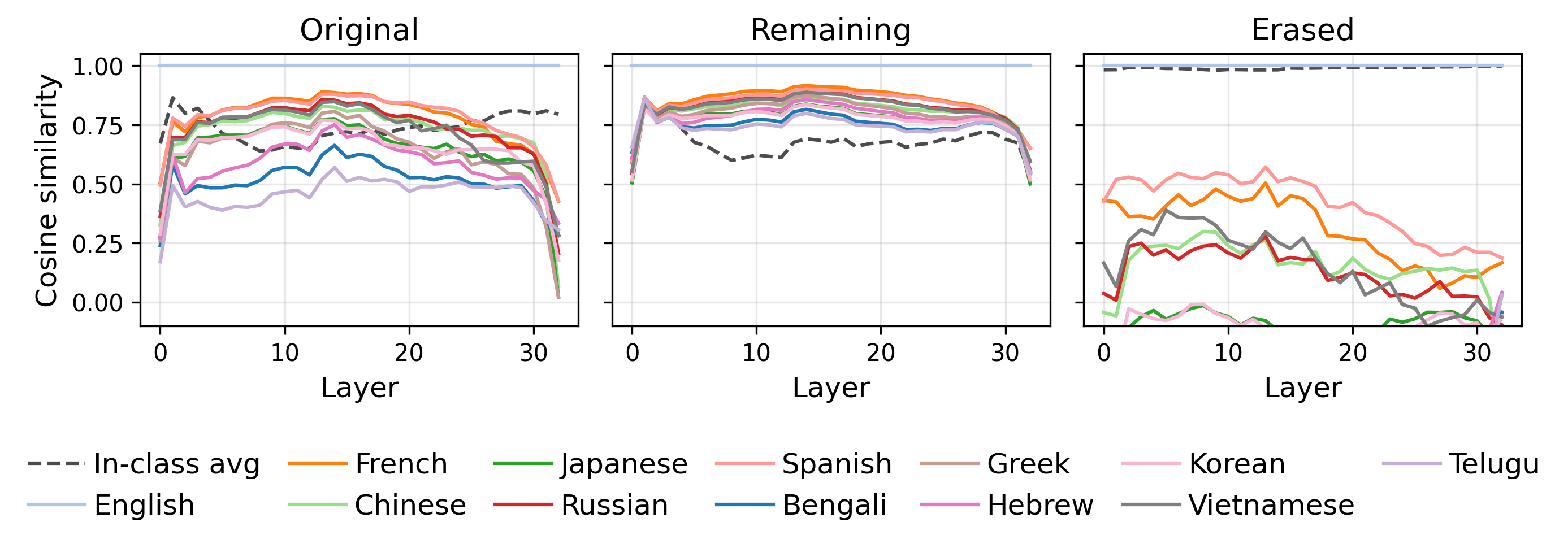}
    \caption{Cosine similarity between the intra-language (dashed, averaged across languages) and inter-language (solid) sentence pairs in the original, remaining, and LEACE-erased representations of Llama3.1-8B across layers.}
    \label{fig:llama_leace_embedding_similarity}
\end{figure*}

%% file: figures/sources/steering/table_steering_results.tex
\begin{table*}[t]
\centering
\small
\setlength{\tabcolsep}{6pt}

\begin{tabular}{@{}ll cc cc cc cc@{}}
\toprule
& & \multicolumn{2}{c}{GMMLU (9)} & \multicolumn{2}{c}{KLAR (7)} & \multicolumn{2}{c}{BELEBELE (9)} & \multicolumn{2}{c}{XQuAD (5)} \\
\cmidrule(lr){3-4} \cmidrule(lr){5-6} \cmidrule(lr){7-8} \cmidrule(lr){9-10}
Model & Method & Acc & Fid & Acc & Fid & Acc & Fid & Acc & Fid \\
\midrule
\multirow{4}{*}{\textit{Llama3.1-8B}}
 & Unsteered          & 48.4 & 93.5 & 68.4 & 94.7 & 74.7 & 87.7 & 35.0 & 77.9 \\
 & Lang Vec           & 43.8 & 93.0 & 69.6\textsuperscript{\dag} & 94.5 & 74.5 & 80.7 & 37.2\textsuperscript{\dag} & 76.8 \\
 & MLRS               & 43.7 & 93.8 & 69.5\textsuperscript{\dag} & 94.6 & 77.5\textsuperscript{\dag} & 88.8 & 35.6\textsuperscript{\dag} & 78.0 \\
 & \OurMethod (Ours)  & \textbf{49.8}\textsuperscript{\dag} & 93.1 & \textbf{73.3}\textsuperscript{\dag} & 93.8 & \textbf{81.9}\textsuperscript{\dag} & 85.3 & \textbf{41.2}\textsuperscript{\dag} & 72.7 \\
\midrule
\multirow{4}{*}{\textit{gemma3-4b}}
 & Unsteered          & 49.4 & 64.7 & 67.5 & 92.7 & 64.2 & 94.4 & 21.9 & 85.9 \\
 & Lang Vec           & 49.0 & 63.9 & 67.9 & 92.8 & 63.8 & 94.1 & 21.8 & 86.3 \\
 & MLRS               & 48.2 & 66.8 & 67.2 & 92.0 & 64.8 & 95.1 & 23.5\textsuperscript{\dag} & 85.1 \\
 & \OurMethod (Ours)  & \textbf{51.8}\textsuperscript{\dag} & 62.6 & \textbf{75.8}\textsuperscript{\dag} & 96.4 & \textbf{66.0}\textsuperscript{\dag} & 92.9 & \textbf{26.0}\textsuperscript{\dag} & 82.5 \\
\midrule
\multirow{4}{*}{\textit{Qwen3-4B}}
 & Unsteered          & 56.8 & 98.5 & 56.8 & 93.7 & 62.3 & 90.1 & 45.5 & 72.0 \\
 & Lang Vec           & 54.9 & 97.0 & 59.1\textsuperscript{\dag} & 93.9 & 63.3\textsuperscript{\dag} & 75.8 & \textbf{45.8} & 68.9 \\
 & MLRS               & 55.9 & 98.5 & 59.2\textsuperscript{\dag} & 93.9 & \textbf{64.9}\textsuperscript{\dag} & 89.2 & 45.6 & 71.6 \\
 & \OurMethod (Ours)  & \textbf{59.6}\textsuperscript{\dag} & 98.1 & \textbf{68.0}\textsuperscript{\dag} & 90.4 & 62.5 & 89.5 & 45.6 & 69.9 \\
\bottomrule
\end{tabular}
\caption{Steering results averaged across target languages.
Each row reports task accuracy and fidelity (\%) of four methods: \textbf{Unsteered}, \textbf{Lang Vec}, \textbf{MLRS}, and \textbf{\OurMethod} (Ours).
The source language is English for all cases, and the number of target languages is shown in parentheses for each dataset.
Bold marks the highest value per metric group.
\textsuperscript{\dag} indicates statistically significant accuracy improvement over Unsteered ($p < 0.05$, one-sided McNemar test across all languages).}
\label{tab:steering_aggregate}
\end{table*}

%% file: figures/sources/steering/figure_llama_steering_results.tex
\begin{figure}[t]
  \includegraphics[width=\columnwidth]{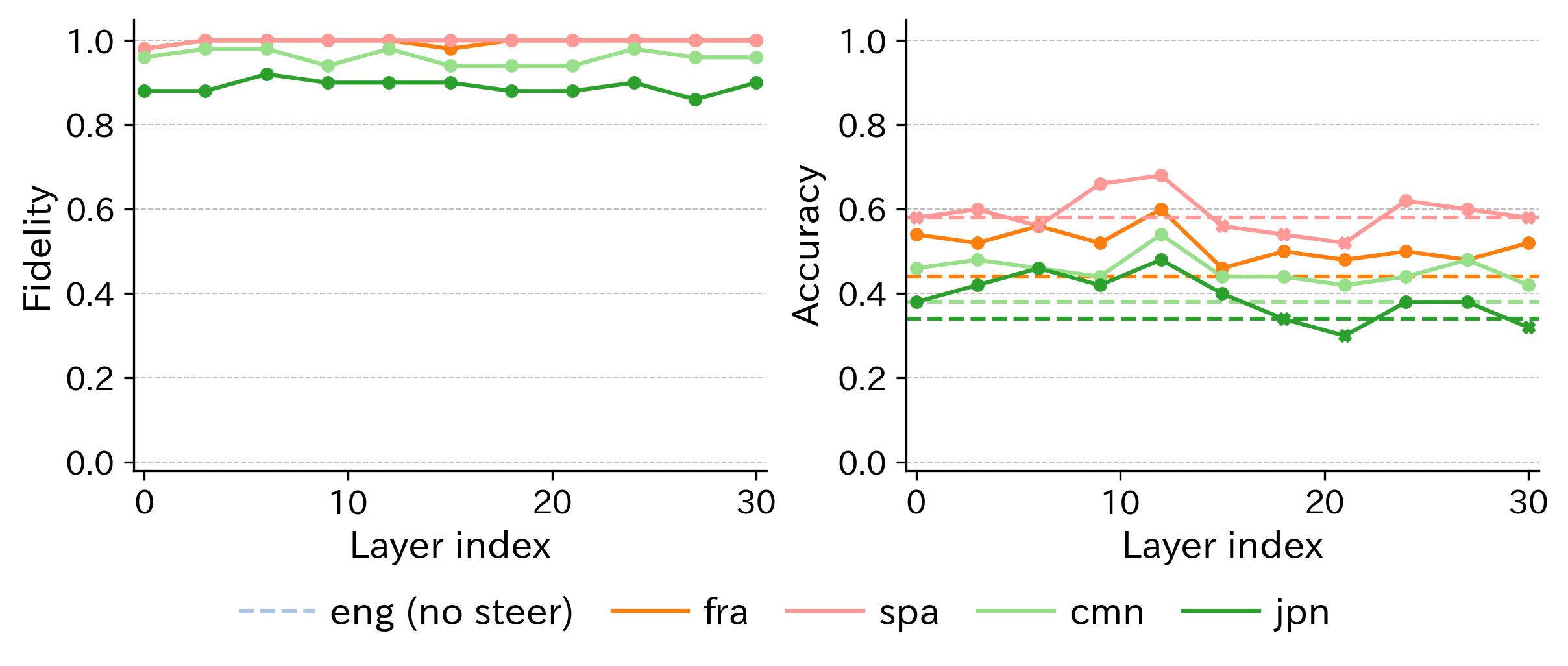}
  \caption{Fidelity (left) and accuracy (right) of Llama3.1-8B on GMMLU dev calculated when conducting hyperparameter search for the best steering layer with \OurMethod.
For other parameters (e.g., $\theta$), we use the best values at each layer.
Dashed lines show the original model (Unsteered) performance.}
  \label{fig:llama_steering_results_gmmlu}
\end{figure}

%% file: figures/sources/steering/figure_llama_semantic_cossim.tex
\begin{figure}[t]
  \includegraphics[width=\columnwidth]{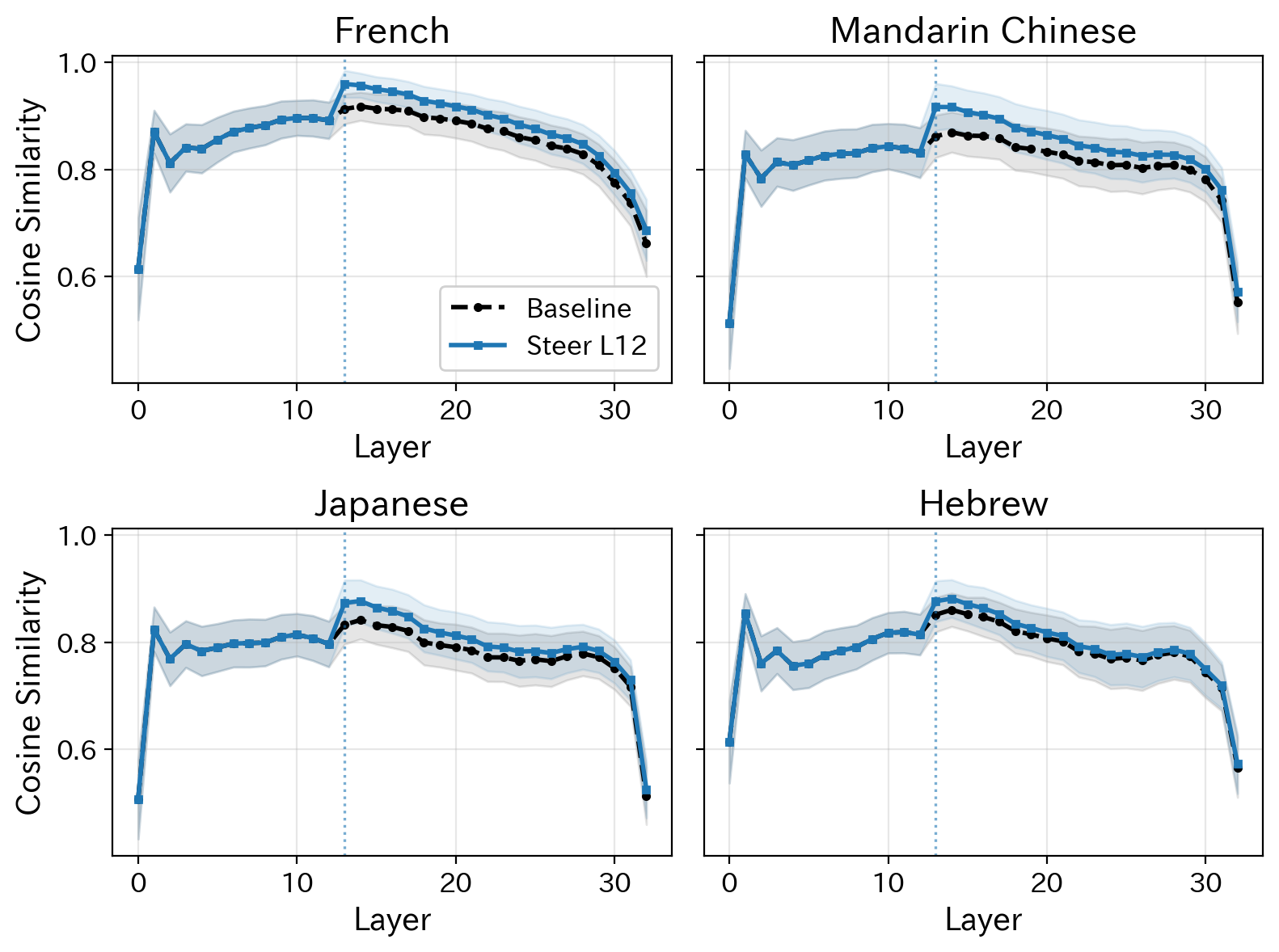}
  \caption{Cosine similarity of the language-agnostic components of sentence representations between the source and target languages when applying \OurMethod steering at layer 12 of Llama3.1-8B.}
  \label{fig:llama_semantic_cossim_alignment_steering}
\end{figure}

%% file: chapters/05_training.tex
\section{Training Experiments}
\label{sec:training}

\input{figures/sources/training/table_train_results.tex}

Finally, we evaluate the training application of \OurMethod described in \S\ref{subsec:using_repr}, where the model is trained to produce language-agnostic representations that match those calculated by \OurMethod.

\subsection{Training Settings}
\label{subsec:training_settings}

To assess the effectiveness of \OurMethod in post-training, we apply instruction-tuning-style training to the base (i.e., non-instruction-tuned) models of Llama3.1-8B, gemma3-4b, and Qwen3-4B, incorporating cross-lingual alignment objectives.
For the alignment data to calculate $\mathcal{L}_\text{align}$, we use WMT24++ and NTREX.
For the instruction-tuning data to calculate $\mathcal{L}_\text{LM}$, we use Aya Dataset~\citep{singh-etal-2024-aya}, a multilingual instruction-tuning dataset.
We train the models in five high-resource languages and evaluate in both ID and OOD languages, including low-resource languages (\autoref{tab:languages}), reflecting the practical scenario where massive training data is available in only high-resource languages.
The evaluation datasets are the same as described in \S\ref{sec:eval_settings}, except that we omit XQuAD.
Unlike the other three datasets, XQuAD requires extractive generation, and the models instruction-tuned in our setting fail to consistently produce answers in the required format under all settings, making the resulting accuracy uninformative.

We compare \OurMethod with standard SFT (\textit{LM}) and \textit{MidAlign}~\citep{liu-niehues-2025-middle}, which minimizes the contrastive loss to increase the similarity of the middle-layer representations on the parallel sentences across languages.
We fine-tune the models with LoRA~\citep{hu2022lora} in all settings, and further details are described in \autoref{app:training_details}.

\subsection{Results and Analysis}
\label{subsec:training_results}

\autoref{tab:train_results} shows the evaluation results.
\OurMethod achieves the best accuracy in 11 out of 18 model-task-language settings (3 models $\times$ 3 tasks $\times$ ID/OOD languages), indicating that it is effective in improving multilingual performance when incorporated into post-training.
To analyze the effect of our training method on the internal representations, we calculate the cosine similarity between the source and target language-agnostic sentence representations before and after training on FLORES+ \textit{devtest} set.
The results for Llama3.1-8B are shown in \autoref{fig:llama_training_cossim} (\autoref{app:training_details}).
We confirm that language-agnostic representations are more aligned after training with \OurMethod, indicating the intended effect of our training objective.

However, the overall performance gains by training are smaller than those by steering (\S\ref{sec:steering_results}).
One possible reason is that the LEACE erasers and covariance matrices are precomputed and fixed during training, but the model's representations change, which may reduce the effectiveness of the alignment loss.
To quantify this effect, we refit LEACE and the covariance matrix on the post-training representations of Llama3.1-8B at layer 12, which is the layer where the alignment loss is applied, and measure principal angles between the subspaces spanned by the top 50 eigenvectors of the pre- and post-training matrices.
The averaged principal angles are 40.3\textdegree{} (max.\ 75.1\textdegree{}) for LEACE and 32.8\textdegree{} (max.\ 87.1\textdegree{}) for the covariance matrix.
These large angles indicate a representation drift, which may make the precomputed parameters increasingly inaccurate.
One possible remedy is periodically updating the LEACE and covariance matrices during training, which we leave for future work.

Moreover, our training with CAROT yields smaller improvements in GMMLU, which includes reasoning-heavy problems.
Although our experiments only consider SFT-based objectives, recent training approaches often adopt reinforcement learning (RL) to improve reasoning capabilities, such as GRPO~\citep{shao2024deepseekmathpushinglimitsmathematical}.
Incorporating CAROT into RL-based training is an interesting direction for future work.

%% file: figures/sources/training/table_train_results.tex
\begin{table*}[t]
\centering\footnotesize
\setlength{\tabcolsep}{3pt}
\begin{tabular}{@{}llcccccc@{}}
\toprule
& & \multicolumn{2}{c}{GMMLU} & \multicolumn{2}{c}{KLAR} & \multicolumn{2}{c}{BELEBELE} \\
\cmidrule(lr){3-4} \cmidrule(lr){5-6} \cmidrule(lr){7-8}
Model & Method & ID & OOD & ID & OOD & ID & OOD \\
\midrule
\multirow{3}{*}{\textit{Llama3.1-8B}}
 & LM                & 43.2 $\pm$ 6.3 & 31.4 $\pm$ 5.6 & 61.7 $\pm$ 11.8 & 54.3 $\pm$ 13.1 & 61.1 $\pm$ 15.4 & 46.3 $\pm$ 5.7 \\
 & MidAlign          & 45.3 $\pm$ 4.9 & \textbf{34.5} $\pm$ 5.9 & 63.1 $\pm$ 13.8 & 47.0 $\pm$ 15.0 & 65.3 $\pm$ 13.2 & 42.3 $\pm$ 9.2 \\
 & \OurMethod (Ours) & \textbf{46.5} $\pm$ 4.8 & 31.8 $\pm$ 5.3 & \textbf{66.9} $\pm$ 9.4 & \textbf{56.1} $\pm$ 12.0 & \textbf{66.5} $\pm$ 15.7 & \textbf{51.5} $\pm$ 8.5 \\
\midrule
\multirow{3}{*}{\textit{gemma3-4b}}
 & LM                & 33.5 $\pm$ 3.2 & 30.5 $\pm$ 2.3 & 54.3 $\pm$ 12.3 & 38.7 $\pm$ 4.2 & 50.3 $\pm$ 6.9 & 40.8 $\pm$ 4.2 \\
 & MidAlign          & \textbf{34.8} $\pm$ 5.5 & \textbf{32.0} $\pm$ 1.4 & 56.4 $\pm$ 15.0 & 41.2 $\pm$ 1.4 & \textbf{60.3} $\pm$ 10.7 & \textbf{50.5} $\pm$ 6.0 \\
 & \OurMethod (Ours) & 34.7 $\pm$ 5.9 & 31.5 $\pm$ 2.7 & \textbf{57.5} $\pm$ 14.1 & \textbf{42.9} $\pm$ 3.8 & 57.8 $\pm$ 8.6 & 49.8 $\pm$ 4.6 \\
\midrule
\multirow{3}{*}{\textit{Qwen3-4B}}
 & LM                & 64.6 $\pm$ 4.5 & 49.4 $\pm$ 5.6 & 71.2 $\pm$ 15.1 & \textbf{51.6} $\pm$ 12.3 & 79.8 $\pm$ 4.6 & \textbf{68.9} $\pm$ 6.5 \\
 & MidAlign          & 65.3 $\pm$ 4.4 & 48.7 $\pm$ 4.1 & 74.6 $\pm$ 13.7 & 40.9 $\pm$ 17.7 & 76.6 $\pm$ 10.2 & 65.7 $\pm$ 4.9 \\
 & \OurMethod (Ours) & \textbf{66.1} $\pm$ 4.0 & \textbf{49.6} $\pm$ 5.8 & \textbf{75.2} $\pm$ 10.5 & 45.0 $\pm$ 15.7 & \textbf{81.9} $\pm$ 8.5 & 68.4 $\pm$ 6.0 \\
\bottomrule
\end{tabular}
\caption{Accuracy (\%) on GMMLU, KLAR, and BELEBELE, aggregating ID and OOD languages.
Values are mean $\pm$ standard deviation across languages within each group.
Here ID/OOD indicates whether the language is included in the fine-tuning data.
Bold marks the highest mean within each ID/OOD group.}
\label{tab:train_results}
\end{table*}

%% file: chapters/06_conclusion.tex
\section{Conclusion}
\label{sec:conclusion}

We present \OurMethod, which disentangles language-specific and language-agnostic components in LLMs' hidden states and aligns language-agnostic representations across languages at the token level while preserving language-specific representations.
Our steering experiments show that representations calculated by \OurMethod enhance the multilingual performance of LLMs without compromising input-output language consistency.
As steering requires a source-language translation of the input and is thus impractical to apply at scale, we propose to use representations obtained by \OurMethod as training objectives, so that translation is needed only during training.
We demonstrate that models trained with \OurMethod outperform previous CLA methods in 11 of 18 evaluation settings, although the gains do not fully match those observed with steering.
Our work provides insights into what constitutes effective alignment targets for CLA in LLMs.

%% file: chapters/appendix.tex
\section{Details on Models and Datasets}
\label{app:models_datasets}

All the model weights used in our experiments are obtained via Hugging Face Hub, and the datasets are obtained from Hugging Face Hub or the official websites of the datasets (e.g., GitHub repositories).
In the steering experiment (\S\ref{sec:steering_results}), we use three instruction-tuned multilingual LLMs: \textit{Llama3.1-8B}\footnote{\url{https://huggingface.co/meta-llama/Llama-3.1-8B-Instruct}}, \textit{gemma3-4b}\footnote{\url{https://huggingface.co/google/gemma-3-4b-it}}, and \textit{Qwen3-4B}\footnote{\url{https://huggingface.co/Qwen/Qwen3-4B}}.
In the training experiment (\S\ref{sec:training}), we use the base (i.e., non-instruction-tuned) variants of the three models.
In \S\ref{sec:eval_settings}, we train LEACE using parallel sentences from two datasets: WMT24++\footnote{\url{https://huggingface.co/datasets/google/wmt24pp}} and NTREX\footnote{\url{https://github.com/MicrosoftTranslator/NTREX/tree/468c6b69c7f6a75d31d4743d9daba2af566cc18d}}.
We also use FLORES+\footnote{\url{https://huggingface.co/datasets/openlanguagedata/flores_plus}} for analysis of the language-specific and language-agnostic representations (\S\ref{sec:eval_lang_repr}).
In \S\ref{sec:steering_results} and \S\ref{sec:training}, we use KLAR Continent category\footnote{\url{https://github.com/boschresearch/KLAR-CLC/tree/02f73c57ea17d4fd9f8b08fd55547058a601ba52}}, Global-MMLU (GMMLU) \textit{test} set\footnote{\url{https://huggingface.co/datasets/CohereLabs/Global-MMLU}}, BELEBELE\footnote{\url{https://huggingface.co/datasets/facebook/belebele}}, and XQuAD\footnote{\url{https://huggingface.co/datasets/google/xquad}} for evaluation.
As an instruction-tuning dataset, we use Aya Dataset\footnote{\url{https://huggingface.co/datasets/CohereLabs/aya_dataset}}.
We list the languages used in each experiment in \autoref{tab:languages}.

\begin{table*}[t]
\small
\centering
\begin{tabular}{ll}
\toprule
\textbf{Split / Benchmark} & \textbf{Languages} \\
\midrule
LEACE ID
  & English, French, Spanish, Mandarin Chinese, Russian, \\
  & Japanese, Greek, Korean, Hebrew, Vietnamese, Bengali, Telugu \\
\midrule
LEACE OOD
  & German, Turkish, Dutch, Ukrainian, Indonesian, Swahili \\
\midrule
Evaluation: KLAR
  & (English), French, Spanish, Mandarin Chinese, Japanese,\\
  & Korean, Hebrew, Greek, (Vietnamese only in \S\ref{sec:training}) \\
\midrule
Evaluation: GMMLU/BELEBELE
  & (English), French, Spanish, Mandarin Chinese, Japanese, Korean, \\
  & Hebrew, Vietnamese, Bengali, Telugu \\
\midrule
Evaluation: XQuAD
  & (English), Spanish, Mandarin Chinese, Vietnamese, Russian, Greek \\
\midrule
Training: ID
  & English, French, Spanish, Mandarin Chinese, Japanese \\
\midrule
Training: OOD
  & (KLAR) Korean, Hebrew, Greek, Vietnamese \\
  & (GMMLU/BELEBELE) Korean, Hebrew, Vietnamese, Bengali, Telugu \\
\midrule
Parameter Search: GMMLU \textit{dev}
  & (English), French, Spanish, Mandarin Chinese, Japanese \\
\bottomrule
\end{tabular}
\caption{Languages used in each experimental split and evaluation benchmark.}
\label{tab:languages}
\end{table*}

\section{LEACE}
\label{app:leace}

LEACE~\citep{belrose2023leace} is a method for concept erasure that calculates an affine transformation to erase a concept from vectors while preserving the remaining information.
Let $\mathvec{X} \in \mathbb{R}^{n \times d}$ be a matrix of $n$ vectors of dimension $d$ and $\mathvec{Z} \in \mathbb{R}^{n \times k}$ be the one-hot encoded concept labels over $k$ classes.
They first calculate the covariance matrix $\mathbf{\Sigma}_\text{XX}$ and the cross-covariance matrix $\mathbf{\Sigma}_\text{XZ}$.
Here, let $\mathvec{W} = (\mathbf{\Sigma}_\text{XX}^{\frac{1}{2}})^+$ and $\mathvec{P}_{\mathvec{W} \mathvec{\Sigma}_\text{XZ}} = (\mathvec{W} \mathvec{\Sigma}_\text{XZ}) (\mathvec{W} \mathvec{\Sigma}_\text{XZ})^+$.
Then, the LEACE transformation $E$ is defined as:
\begin{equation}
    E(\mathvec{x}) = \mathvec{P} \mathvec{x} + \mathbb{E}[\mathvec{x}] - \mathvec{P} \mathbb{E}[\mathvec{x}],
\end{equation}
where $\mathvec{P} = \mathbf{I} - \mathvec{W}^+ \mathvec{P}_{\mathvec{W} \mathvec{\Sigma}_\text{XZ}} \mathvec{W}$ and $\mathbb{E}[\mathvec{x}]$ is the mean of $\mathvec{X}$.
This transformation ensures that no linear classifier can predict $\mathvec{Z}$ from $E(\mathvec{X})$ better than a constant predictor, while minimizing the expected squared distance between the original and transformed vectors.
We apply LEACE to hidden states $\mathvec{h}$ of LLMs with the language of the input sentence as the concept $\mathvec{Z}$, so that $E(\mathvec{h})$ gives the language-agnostic component and $\mathvec{h}-E(\mathvec{h})$ the language-specific one.

\section{Additional Results for Language-Specific and Language-Agnostic Representations}
\label{app:additional_lang_repr_results}

\input{figures/sources/appendix/figure_llama_leace_scatter.tex}

In \S\ref{sec:eval_lang_repr}, we show the 2D scatter plots of the original, remaining, and LEACE-erased representations of Llama3.1-8B at layer 12.
Here, we show the scatter plots of Llama3.1-8B at layers 6 and 25 in \autoref{fig:llama_leace_scatter_layer006_025}.
We can observe that the same trend as layer 12 holds at both layers, with the LEACE-erased representations showing clearer separation by language, while the remaining representations are mixed together regardless of language.

\input{figures/sources/appendix/figure_gemma_qwen_leace_scatter.tex}

We also show the scatter plots of gemma3-4b and Qwen3-4B at layers 6 and 7 in \autoref{fig:gemma_qwen_leace_scatter_layer006_007}.
For gemma3-4b, the original representations are more mixed across languages in the PCA-based visualization compared to other models.
This can be due to anisotropic distribution of the hidden states of gemma3-4b, with a large contribution ratio of the top principal component (PC) (we observe 89.6\% contribution ratio of the top PC at layer 13).
Nonetheless, LEACE successfully identifies language representations, as shown by the similar trend in the scatter plots.

\input{figures/sources/appendix/figure_gemma_qwen_leace_ap.tex}

Finally, we train a linear classifier to identify the language of the input sentence on the representations on the FLORES+ \textit{dev} set and evaluate classification accuracy on the FLORES+ \textit{devtest} set.
We show the language classification accuracy before and after LEACE transformation for the three models in \autoref{fig:llama_gemma_qwen_leace_ap}.
After applying LEACE, classification accuracy for ID languages drops clearly, indicating successful erasure of linearly encoded language-specific information.
Although the accuracy remains around 50--80\%, this can be because the training and evaluation datasets are from the same domain (i.e., FLORES+) while LEACE is trained on a different domain (i.e., WMT24++ and NTREX).
Also, as the number of samples for training the classifier is limited, the classifier may overfit.
For OOD languages, the classifier retains almost 100\% accuracy even after applying LEACE, indicating residual language-specific information.

\section{Optimal Transport Plans}
\label{app:ot_plans}

\input{figures/sources/appendix/figure_llama_ot_map.tex}

\autoref{fig:llama_ot_map} shows examples of the transport plans calculated in \OurMethod for Llama3.1-8B.
The transport plans are sparse, as a result of the perplexity-based filtering (\autoref{eq:perplexity}), which can increase robustness of alignment by suppressing noise.
We can also see that semantically corresponding tokens across languages are aligned (e.g., ``(Tur)quie'' and ``Turkey''), which supports the effectiveness of \OurMethod in aligning semantic representations.

\section{Evaluation Protocols}
\label{app:eval_protocols}

This section describes the evaluation configurations for the steering and training experiments in \S\ref{sec:steering_results} and \S\ref{sec:training}.
We use four datasets for evaluation: KLAR Continent, GMMLU, BELEBELE, and XQuAD.
We select these datasets because they provide parallel translations and are neutral in terms of language and culture, allowing fair evaluation of multilingual performance.
For KLAR, five prompt templates in each language are provided in the official repository, and we use all of them for evaluation, resulting in 1,060 samples per language.
For GMMLU, BELEBELE, and XQuAD, we prepare one prompt template in English and translate it into other languages with ChatGPT\footnote{\url{https://chatgpt.com/overview/}}.
We randomly sample 30 instances for each GMMLU category and language to reduce the computational cost, resulting in 1,710 samples per language.
BELEBELE has 900 samples per language, and XQuAD has 1,190 samples per language.
For KLAR, GMMLU, and BELEBELE, we use \textit{test} sets for evaluation, while for XQuAD, we use the \textit{validation} set, since they only provide the \textit{validation} set.

We let the model generate the answer with greedy decoding up to 50 tokens for KLAR and XQuAD, 500 tokens for GMMLU, and 100 tokens for BELEBELE, expecting that the reasoning tasks will require more tokens.
We extract the final answer from the generated text in a rule-based manner based on the instructions in the prompt templates.

For hyperparameter tuning, we use the \textit{dev} set of GMMLU (randomly sampled 50 instances per language) and perform a grid search.
Regarding the parameters for Sinkhorn algorithm to solve optimal transport in \OurMethod (\autoref{eq:sinkhorn}), we set the regularization parameter $\epsilon_{\mathrm{OT}}$ to 0.05 and the unbalanced mass penalty $\rho$ to 0.5.

\section{Baseline Steering Methods}
\label{app:baseline_steering}

In the steering experiments (\S\ref{sec:steering_results}), we compare \OurMethod with two baseline methods: \textit{Lang Vec} and \textit{MLRS}.

\textit{Lang Vec} adds the difference between the mean hidden vectors of the source and target languages to the hidden states at a specific layer to bring the representation of the target language closer to that of the source language:
\begin{equation}
    \mathvec{h}_\text{tgt}^{\,\text{new}} = \mathvec{h}_\text{tgt} + \alpha (\bar{\mathvec{h}}_\text{src} - \bar{\mathvec{h}}_\text{tgt}).
\end{equation}
We calculate the mean vectors on the sentence representations (\autoref{eq:sentence_repr}) of the same parallel datasets as used for training LEACE (\S\ref{sec:eval_settings}).
$\alpha$ is a hyperparameter that controls the strength of the steering.
We search $\alpha$ from \{0.5, 1.0, 1.5, 2.0\} and the layer from every three layers for each model.

For \textit{MLRS}~\citep{zhao2026when}, it first calculates the mean hidden vectors of the last token activations of each language in the training data and concatenates them to form $\mathvec{M} \in \mathbb{R}^{d \times L}$, where $L$ is the number of languages.
Then it minimizes the following objective:
\begin{align}
    \min_{\mathvec{M}_a, \mathvec{M}_s, \Gamma} \| \mathvec{M} - \mathvec{M}_a \mathvec{1}^\top - \mathvec{M}_s \Gamma \|_F^2 \\
    \text{s.t. } \text{Span}(\mathvec{M}_a) \perp \text{Span}(\mathvec{M}_s),
\end{align}
where $\mathvec{M}_a \in \mathbb{R}^{d \times 1}$ is the language-agnostic component, $\mathvec{M}_s \in \mathbb{R}^{d \times r}$ is the language-specific component, and $\Gamma \in \mathbb{R}^{r \times L}$ is the coefficient matrix.
Following the paper that MLRS is based on~\citep{xie-etal-2022-discovering}, we set $r=L-1$.
The optimal solution of this objective can be obtained using singular value decomposition.
At inference time, MLRS intervenes in the language-specific component of the hidden states as follows:
\begin{equation}
    \mathvec{h}_\text{tgt}^{\,\text{new}} = \mathvec{h}_\text{tgt} - \lambda \mathvec{M}_s \mathvec{M}_s^\top \mathvec{h}_\text{tgt},
\end{equation}
where $\lambda$ is a hyperparameter that controls the direction and strength of the intervention.
We follow the original paper and set $\lambda>0$ (suppressing language-specific information) from the one-third to two-thirds layers and $\lambda<0$ (adding back language-specific information) for the last one-third layers.
We search $\lambda_{\text{sup}}$ from \{0.1, 0.2, 0.3, 0.4\} for suppressing and $\lambda_{\text{add}}$ from \{-0.1, -0.2, -0.3, -0.4\} for adding back.

\section{Selected Hyperparameters}
\label{app:selected_parameter}

\begin{table*}[t]
  \small
  \centering
  \begin{tabular}{lccc}
    \toprule
    \textbf{Method} & \textbf{Llama3.1-8B} & \textbf{gemma3-4b} & \textbf{Qwen3-4B} \\
    \midrule
    \OurMethod & \{layer: 12, $\theta$: 0.25\} & \{layer: 6, $\theta$: 0.5\} & \{layer: 9, $\theta$: 0.75\} \\
    Lang Vec & \{layer: 21, $\alpha$: 0.5\} & \{layer: 33, $\alpha$: 0.5\} & \{layer: 27, $\alpha$: 2.0\} \\
    MLRS & \{$\lambda_{\text{sup}}$: 0.1, $\lambda_{\text{add}}$: -0.4\} & \{$\lambda_{\text{sup}}$: 0.1, $\lambda_{\text{add}}$: -0.4\} & \{$\lambda_{\text{sup}}$: 0.1, $\lambda_{\text{add}}$: -0.4\} \\
    \bottomrule
  \end{tabular}
  \caption{Selected hyperparameters for each steering method and model.}
  \label{tab:steering_hyperparameters}
\end{table*}

\begin{table*}[t]
  \small
  \centering
  \begin{tabular}{lccc}
    \toprule
    \textbf{Method} & \textbf{Llama3.1-8B} & \textbf{gemma3-4b} & \textbf{Qwen3-4B} \\
    \midrule
    \OurMethod & \{lr: 5e-5\} & \{lr: 1e-4\} & \{lr: 1e-4\} \\
    LM & \{lr: 1e-5\} & \{lr: 1e-5\} & \{lr: 5e-5\} \\
    MidAlign & \{lr: 5e-5, $\tau$: 0.1\} & \{lr: 1e-4, $\tau$: 0.1\} & \{lr: 1e-4, $\tau$: 0.1\} \\
    \bottomrule
  \end{tabular}
  \caption{Selected hyperparameters for each training method and model.}
  \label{tab:training_hyperparameters}
\end{table*}

In the steering experiments (\S\ref{sec:steering_results}) and training experiments (\S\ref{sec:training}), we select the best hyperparameters for each method based on the evaluation accuracy on the GMMLU \textit{dev} set.
The selected hyperparameters for steering methods are shown in \autoref{tab:steering_hyperparameters}, and those for training methods are shown in \autoref{tab:training_hyperparameters}.
Note that for the layer and $\theta$ of \OurMethod in the training experiments, we use the best values selected in the steering experiments.

\section{Additional Steering Results}
\label{app:additional_steering_results}

\input{figures/sources/appendix/table_steering_agreement.tex}

\paragraph{Agreement Ratio of Steering Results.}

We show the agreement ratio (i.e., the percentage of instances for which the model's answer in the target language matches that in the source language) of each steering method in \autoref{tab:agreement}, along with the accuracy reported in \autoref{tab:steering_aggregate}.
The agreement ratio directly measures CLA, and a higher agreement ratio indicates that the model is more consistent in its answers across languages.
Overall, the agreement ratio correlates with the accuracy, and \OurMethod achieves the highest agreement ratio in most cases.
Moreover, \OurMethod tends to increase the agreement ratio more than the accuracy (e.g., for Llama3.1-8B on GMMLU, the accuracy increases by 1.4 points while the agreement ratio increases by 2.4 points).
This indicates that the increase in the accuracy of \OurMethod is due to the improvement in CLA.

\input{figures/sources/appendix/figure_steering_results_gmmlu.tex}

\paragraph{Detailed Layer-wise and Language-wise Steering Results.}

We present the layer-wise per-language accuracy and fidelity when applying steering with \OurMethod, calculated during the hyperparameter search on GMMLU dev.
The results for Llama3.1-8B are shown in \autoref{fig:llama_steering_results_gmmlu}, and the results for gemma3-4b and Qwen3-4B are shown in \autoref{fig:gemma_qwen_steering_results_gmmlu}.
\OurMethod improves accuracy across languages, especially in the early layers, and maintains high fidelity in most cases.
However, for gemma3-4b, the fidelity is low in some languages (e.g., Chinese and Japanese).
This can be because gemma3-4b originally (without steering) tends to generate English reasoning steps even when prompted in other languages.

\input{figures/sources/appendix/figure_langvec_steering_results_gmmlu_dev.tex}

Furthermore, we show the layer-wise and per-language results of steering with Lang Vec on GMMLU dev in \autoref{fig:llama_gemma_qwen_steering_results_lang_vec_gmmlu_dev}.
The accuracy is less improved relative to \OurMethod, and the fidelity often becomes low in early-to-middle layers.
Since Lang Vec adds the diff-in-means language vector to the hidden states of the target language, it can remove language information and thus change the output language, which can cause low fidelity.
On the other hand, \OurMethod explicitly preserves the language representations, which can contribute to high fidelity.

\paragraph{Ablation of Each Component of \OurMethod.}

\input{figures/sources/appendix/table_ablation_steering.tex}

We conduct an ablation study to analyze the contribution of each component of \OurMethod, including LEACE-based decomposition, perplexity-based filtering of the OT plan, and the token-level OT alignment.
\autoref{tab:steering_ablation} compares the evaluation accuracy and the fidelity of \OurMethod, \OurMethod without LEACE-based decomposition (\textit{No LEACE}), \OurMethod without filtering the OT plan (\textit{No Filter}), and swapping only the language-agnostic representations of the last token between the inputs in source and target languages (\textit{Last Token}).
Last Token corresponds to \OurMethod without token-level OT alignment.
Note that No Filter is included in the hyperparameter search for \OurMethod, which corresponds to $\theta=1.0$.

The results show that \OurMethod is most stable across models and benchmarks.
\OurMethod outperforms No LEACE and Last Token in 9 and 10 out of 12 cases, respectively, in terms of accuracy.
Although the accuracies of No Filter are comparable to those of \OurMethod, No Filter is sometimes unstable (e.g., for Qwen3-4B on KLAR, the accuracy is 14.0 points lower than \OurMethod), and the fidelity is often lower than \OurMethod (9 out of 12 cases).
These results suggest that LEACE-based decomposition and filtering of the OT plan make \OurMethod more robust, and that token-level OT alignment is effective in improving accuracy.

\section{Training Details}
\label{app:training_details}

\paragraph{MidAlign.}
Here, we provide the detailed specification of the training algorithms for the baseline method, MidAlign.
MidAlign fine-tunes the model by alternating between a contrastive step and a language-modeling (LM) step.
In the contrastive step, at the specified layer $l$, the source and target hidden states are calculated by mean-pooling hidden vectors over non-padding tokens, giving $\mathbf{h}^{(l)}_s, \mathbf{h}^{(l)}_t \in \mathbb{R}^{B \times d}$, where $B$ is the batch size and $d$ is the hidden dimension.
The symmetric InfoNCE loss is then computed as:
\begin{align}
    \mathcal{L}^{(l)}_{\text{NCE, col}} &= -\log \frac{\exp(\mathrm{sim}(\mathbf{h}^{(l)}_s, \mathbf{h}^{(l)}_t) / \tau)}{\sum_{i=1}^B \exp(\mathrm{sim}(\mathbf{h}^{(l)}_s, \mathbf{h}^{(l)}_{t,i}) / \tau)}, \\
    \mathcal{L}^{(l)}_{\text{NCE, row}} &= -\log \frac{\exp(\mathrm{sim}(\mathbf{h}^{(l)}_t, \mathbf{h}^{(l)}_s) / \tau)}{\sum_{i=1}^B \exp(\mathrm{sim}(\mathbf{h}^{(l)}_t, \mathbf{h}^{(l)}_{s,i}) / \tau)}, \\
    \mathcal{L}^{(l)}_{\text{contrastive}} &= \frac{1}{2} (\mathcal{L}^{(l)}_{\text{NCE, col}} + \mathcal{L}^{(l)}_{\text{NCE, row}})
\end{align}
where $\tau$ is the temperature and $\mathrm{sim}(\cdot, \cdot)$ is the cosine similarity function.
As in the original paper, we use the alignment datasets for the contrastive step (parallel dataset) and the task-specific dataset for the LM step.

\paragraph{Training Parameters.}

We train all models for one epoch with a batch size of 16 using the AdamW optimizer.
The learning rate is searched from \{5e-4, 1e-4, 5e-5, 1e-5\} for each model and method.
For MidAlign, we search the temperature $\tau$ from \{0.1, 0.5, 1.0\}.
The selected hyperparameters are shown in \autoref{tab:training_hyperparameters}.
For other parameters, we set weight decay to 0.01 and adopt a linear learning rate scheduler with 300 warmup steps.
We apply LoRA to all linear modules with rank $r=64$, scaling factor $\alpha=64$, and a dropout rate of 0.05.

We limit the maximum number of instruction-tuning training samples to 2,000 per language to balance the number of training data across languages, and we sample the same number of parallel sentences for the alignment step.
We set the maximum input length to 500 tokens for task-specific data and 64 tokens for parallel sentences.
As for the training layers for MidAlign, we select layers 16, 17, and 18 for Llama3.1-8B, gemma3-4b, and Qwen3-4B, respectively, following the analysis in the original paper, which shows that the middle layers are most effective for MidAlign.

\paragraph{Analysis of Training.}

\input{figures/sources/appendix/figure_llama_training_cossim.tex}

To further analyze the training results, we calculate the cosine similarity between the source and target language-agnostic representations at each layer before and after training.
The results for Llama3.1-8B are shown in \autoref{fig:llama_training_cossim}.
We observe that the similarity increases most significantly around the training layer (layer 12 for Llama3.1-8B) after training with \OurMethod, which indicates that the training successfully encourages the model to produce language-agnostic representations that are close to the source representations.
The increase in similarity lasts across layers, which can contribute to the improved performance in the evaluation benchmarks.

\section{Computational Resources}

All the experiments are conducted on a single NVIDIA H100 GPU.
Hyperparameter search of the steering experiments in \S\ref{sec:steering_results} took around 2 hours for each model and method.
Hyperparameter search of the training experiments in \S\ref{sec:training} took around 1.5 hours for each model for LM and \OurMethod, and around 4 hours for MidAlign, since we search the temperature $\tau$ in addition to the learning rate for MidAlign.
For the evaluation of each model and method, it took 2.5 hours for GMMLU, 20 minutes for KLAR, BELEBELE, and XQuAD, respectively.

\section{LLM Usage}

We used Claude Code\footnote{\url{https://code.claude.com/docs/en/overview}} as a coding assistant for implementing the scripts for the experiments and visualizations.
We also used ChatGPT to proofread the paper and to translate the prompt template used in evaluation into different languages.

%% file: figures/sources/appendix/figure_llama_leace_scatter.tex
\begin{figure}[t]
  \centering
  \begin{minipage}{0.95\columnwidth}
    \centering
    \includegraphics[width=\columnwidth]{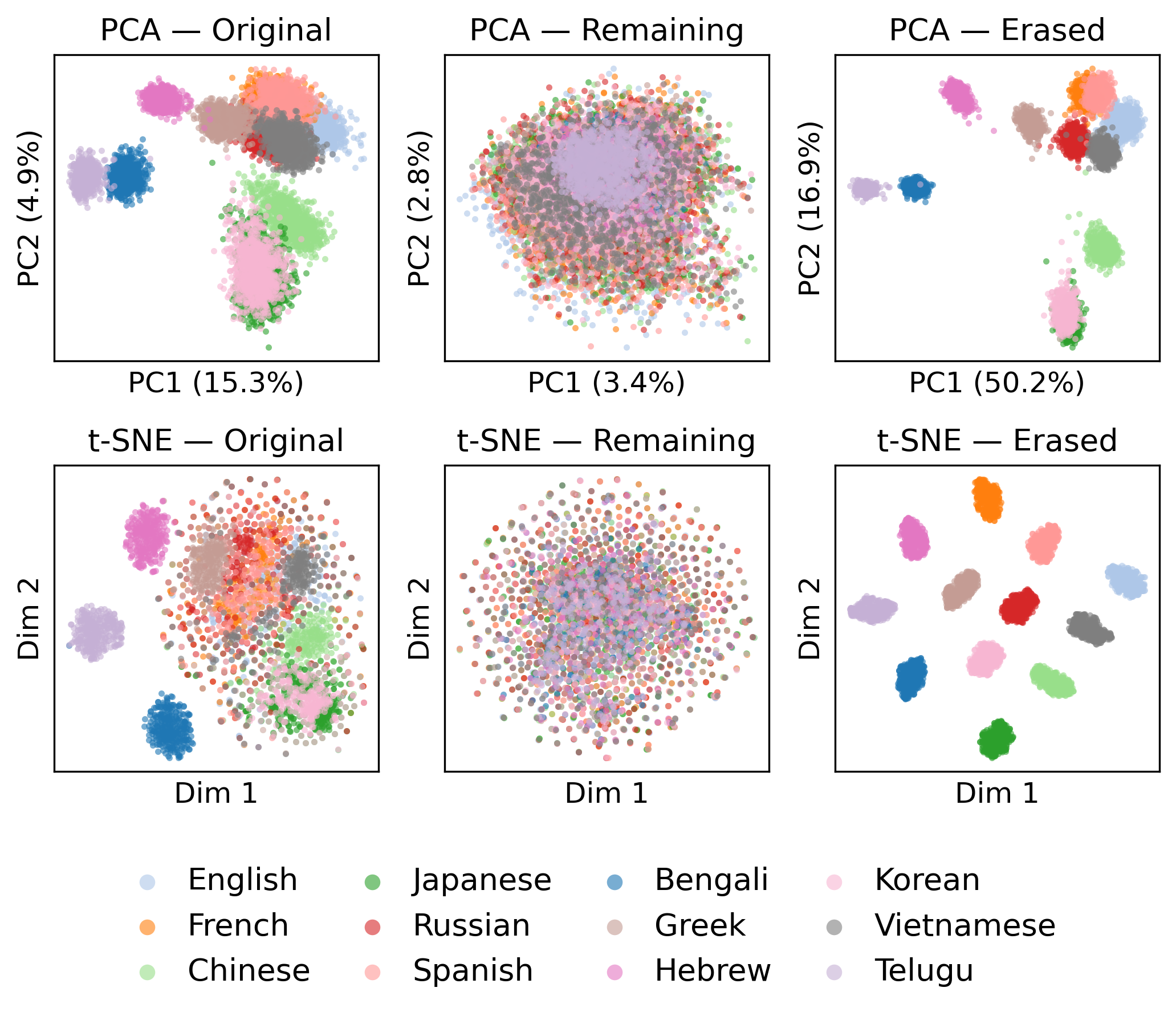}
    \subcaption{Layer 6}
    \label{fig:llama_leace_scatter_layer006_id}
  \end{minipage}
  \hspace{5mm}
  \begin{minipage}{0.95\columnwidth}
    \includegraphics[width=\columnwidth]{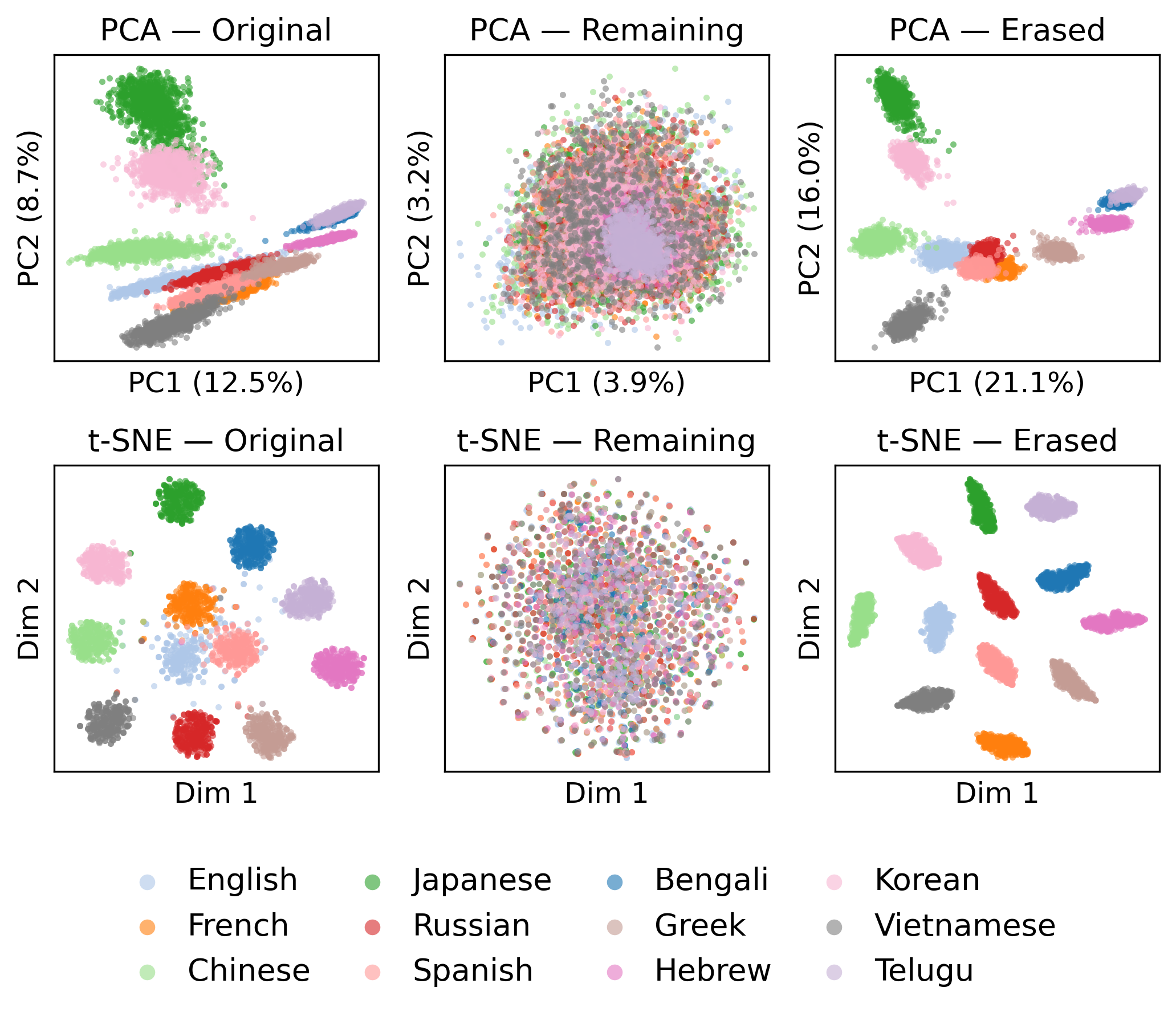}
    \subcaption{Layer 25}
    \label{fig:llama_leace_scatter_layer025_id}
  \end{minipage}
  \caption{Scatter plot of original (left), remaining (center), and LEACE-erased (right) representations of Llama3.1-8B at layer 6 and 25.
  Each point represents the representation of a sentence, colored by language.}
  \label{fig:llama_leace_scatter_layer006_025}
\end{figure}

%% file: figures/sources/appendix/figure_gemma_qwen_leace_scatter.tex
\begin{figure}[t]
  \centering
  \begin{minipage}{0.95\columnwidth}
    \centering
    \includegraphics[width=\columnwidth]{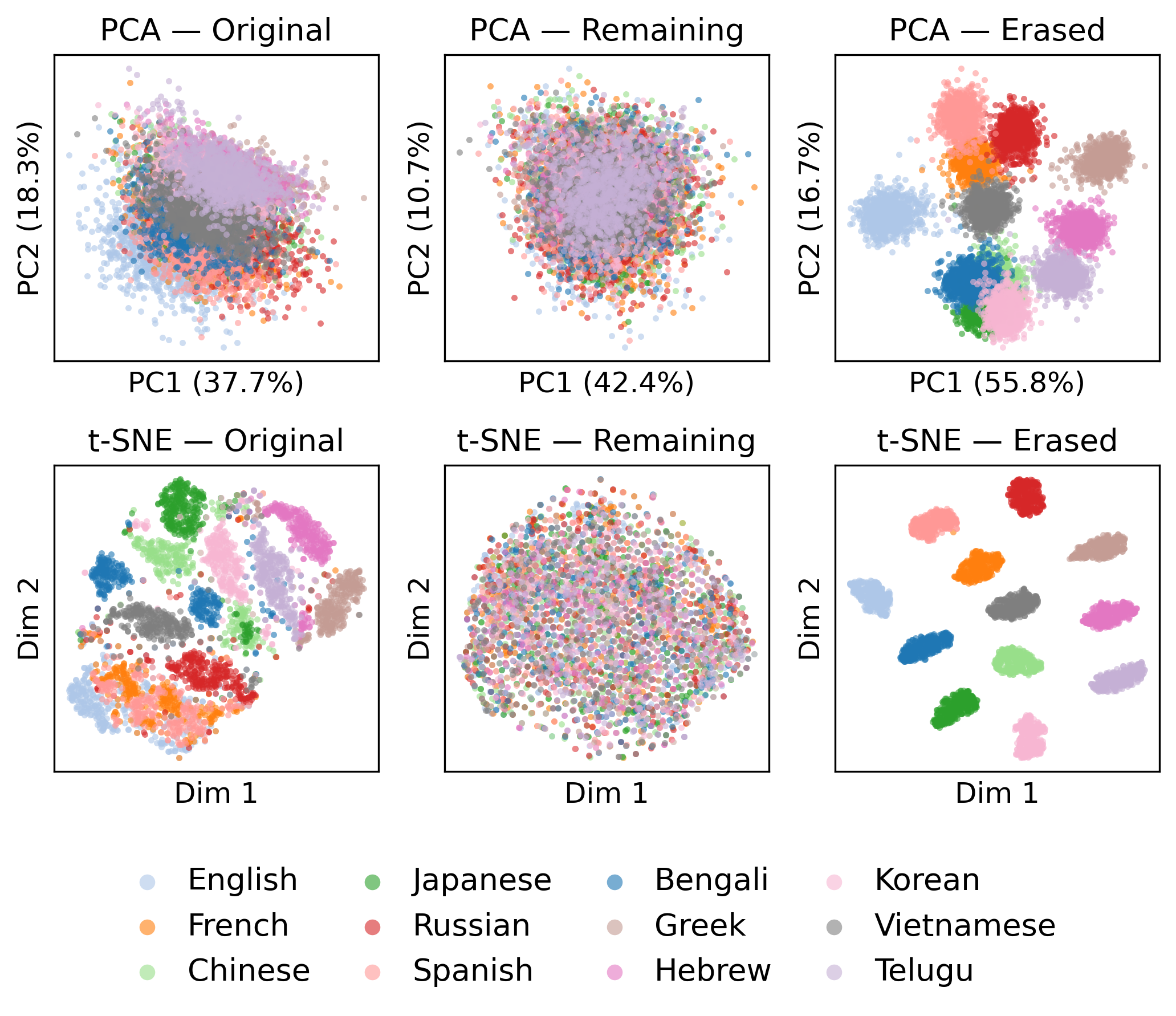}
    \subcaption{gemma3-4b layer 6}
    \label{fig:gemma_leace_scatter_layer006_id}
  \end{minipage}
  \hspace{5mm}
  \begin{minipage}{0.95\columnwidth}
    \includegraphics[width=\columnwidth]{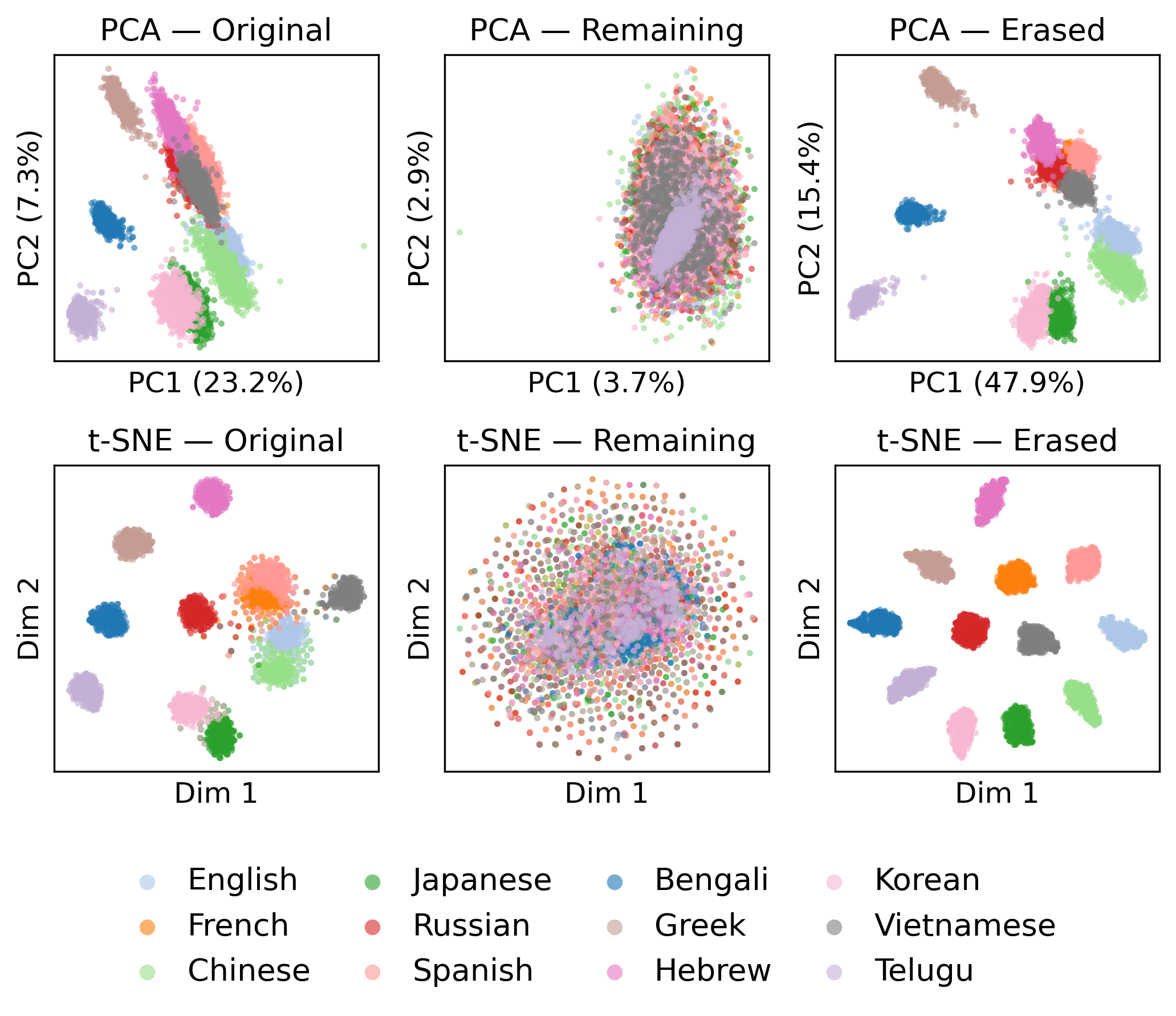}
    \subcaption{Qwen3-4B layer 7}
    \label{fig:qwen_leace_scatter_layer007_id}
  \end{minipage}
  \caption{Scatter plot of original (left), remaining (center), and LEACE-erased (right) representations of gemma3-4b and Qwen3-4B at layers 6 and 7.
  Each point represents the representation of a sentence, colored by language.}
  \label{fig:gemma_qwen_leace_scatter_layer006_007}
\end{figure}

%% file: figures/sources/appendix/figure_gemma_qwen_leace_ap.tex
\begin{figure}[t]
    \centering
    \begin{minipage}{0.85\columnwidth}
        \includegraphics[width=\columnwidth]{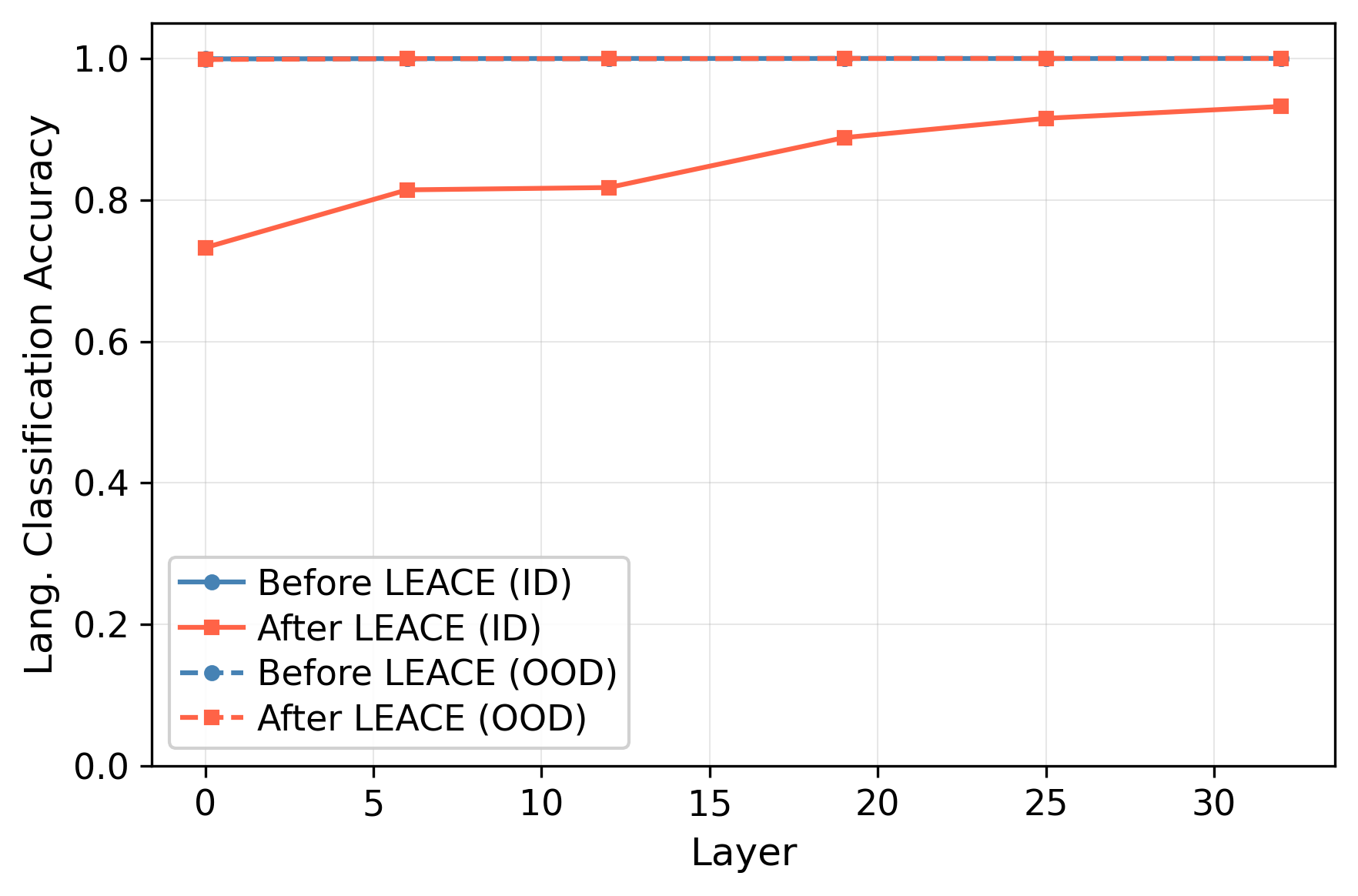}
        \subcaption{Llama3.1-8B}
        \label{fig:llama_leace_ap}
    \end{minipage}
    \begin{minipage}{0.85\columnwidth}
        \includegraphics[width=\columnwidth]{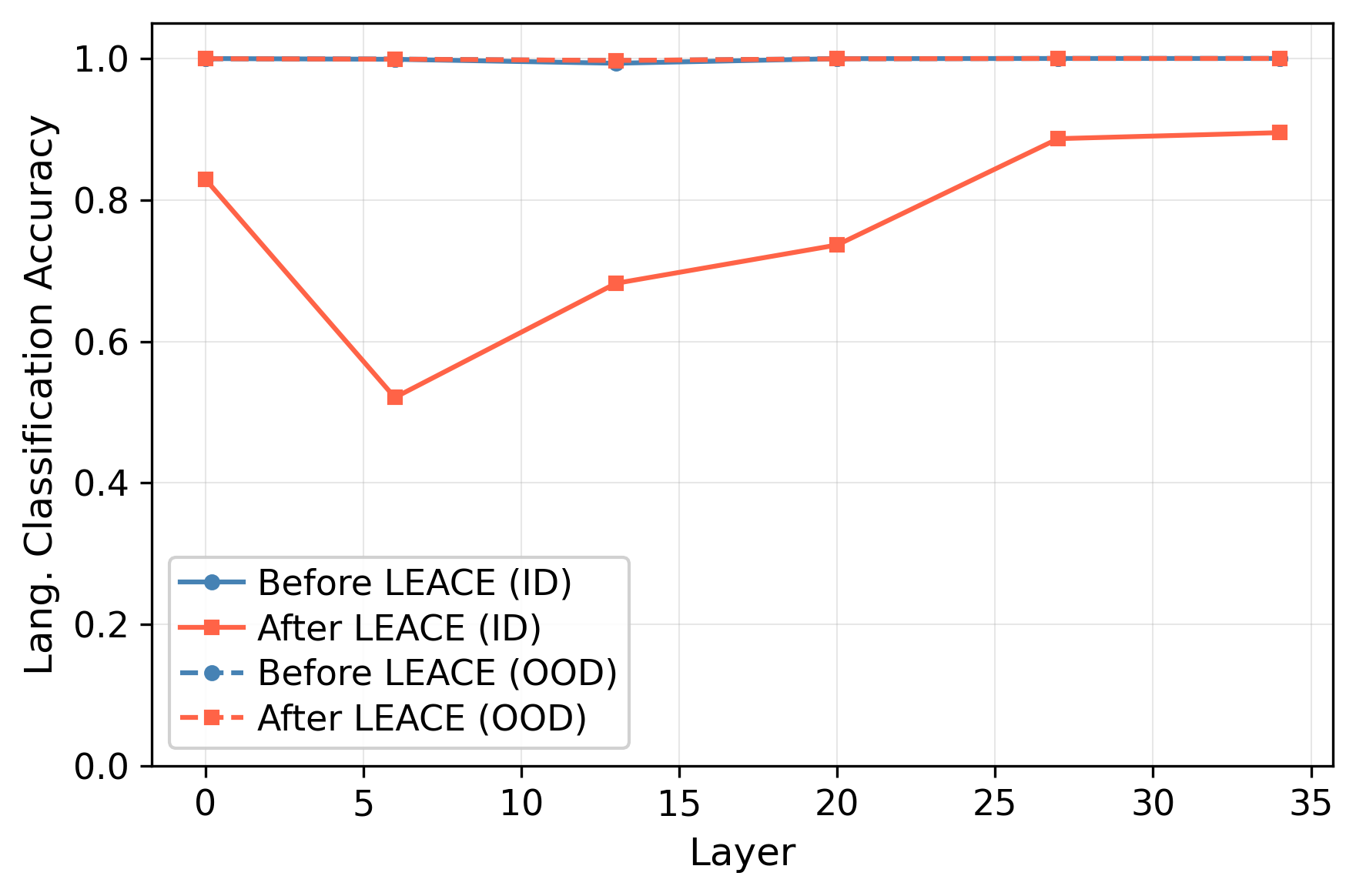}
        \subcaption{gemma3-4b}
        \label{fig:gemma_leace_ap}
    \end{minipage}
    \hspace{5mm}
    \begin{minipage}{0.85\columnwidth}
        \includegraphics[width=\columnwidth]{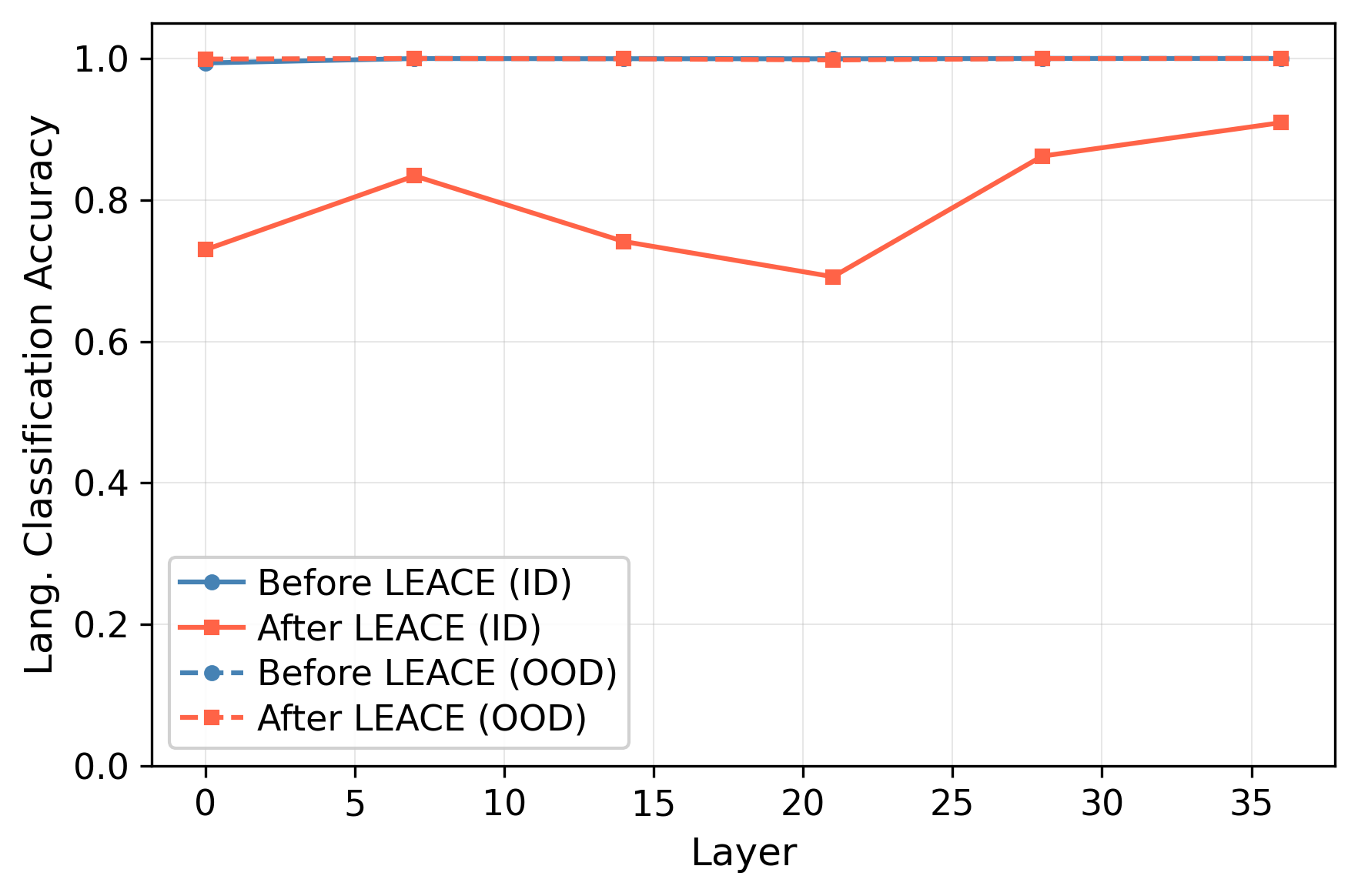}
        \subcaption{Qwen3-4B}
        \label{fig:qwen_leace_ap}
    \end{minipage}
    \caption{Language classification accuracy before (blue) and after (red) LEACE transformation at every six layers of Llama3.1-8B, gemma3-4b and Qwen3-4B.
We report the accuracies of in-distribution (solid) and out-of-distribution (dashed) languages separately.
The accuracies for ID languages before LEACE and for OOD languages before and after LEACE are almost 100\%.}
    \label{fig:llama_gemma_qwen_leace_ap}
\end{figure}

%% file: figures/sources/appendix/figure_llama_ot_map.tex
\begin{figure}[t]
    \centering
    \begin{minipage}{0.85\columnwidth}
        \includegraphics[width=\columnwidth]{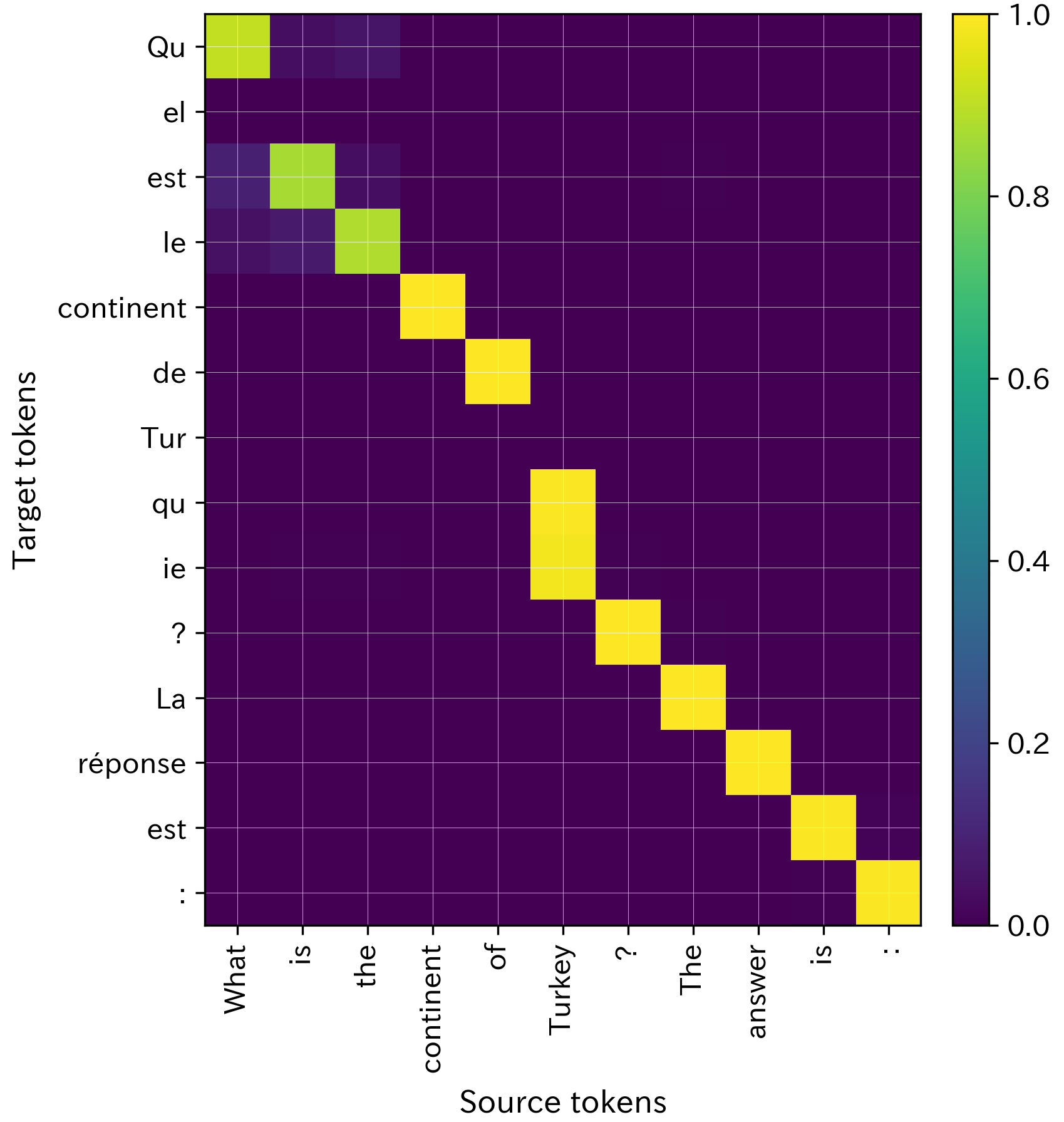}
        \subcaption{French at layer 12}
        \label{fig:llama_ot_map_french}
    \end{minipage}
    \hspace{5mm}
    \begin{minipage}{0.85\columnwidth}
        \includegraphics[width=\columnwidth]{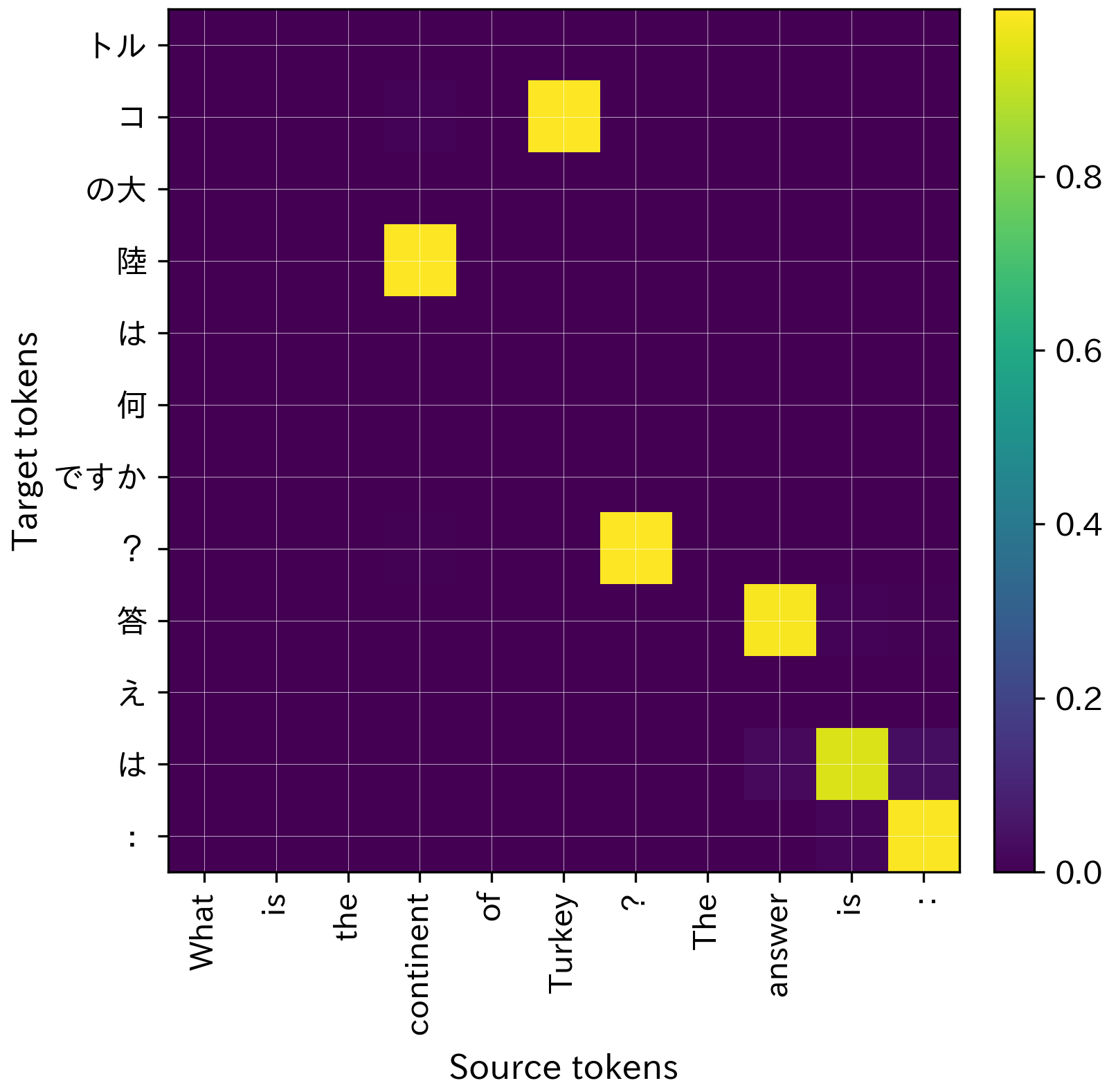}
        \subcaption{Japanese at layer 12}
        \label{fig:llama_ot_map_japanese}
    \end{minipage}
    \caption{Visualization of the calculated OT transport plan $\mathvec{T}$ on Llama3.1-8B for a sample sentence pair in French-English (a) and Japanese-English (b).
The brightness of each cell corresponds to the value of $T_{ij}$, indicating how much the $j$-th token in the source language (English) is transported to the $i$-th token in the target language (French or Japanese).} 
    \label{fig:llama_ot_map}
\end{figure}

%% file: figures/sources/appendix/table_steering_agreement.tex
\begin{table}[t]
\centering\small
\setlength{\tabcolsep}{5pt}
\begin{tabular}{@{}llcccc@{}}
\toprule
& & \multicolumn{2}{c}{GMMLU} & \multicolumn{2}{c}{BELEBELE} \\
\cmidrule(lr){3-4} \cmidrule(lr){5-6}
Model & Method & Acc & Agr & Acc & Agr \\
\midrule
\multirow{4}{*}{\textit{Llama3.1-8B}}
 & Unsteered  & 48.4 & 51.5 & 74.7 & 75.3 \\
 & Lang Vec   & 43.8 & 47.1 & 74.5 & 75.2 \\
 & MLRS       & 43.7 & 47.0 & 77.5 & 78.2 \\
 & \OurMethod & \textbf{49.8} & \textbf{53.9} & \textbf{81.9} & \textbf{83.9} \\
\midrule
\multirow{4}{*}{\textit{gemma3-4b}}
 & Unsteered  & 49.4 & 57.0 & 64.2 & 64.1 \\
 & Lang Vec   & 49.0 & 56.9 & 63.8 & 63.3 \\
 & MLRS       & 48.2 & 56.2 & 64.8 & 63.8 \\
 & \OurMethod & \textbf{51.8} & \textbf{61.4} & \textbf{66.0} & \textbf{65.6} \\
\midrule
\multirow{4}{*}{\textit{Qwen3-4B}}
 & Unsteered  & 56.8 & 61.2 & 62.3 & 62.7 \\
 & Lang Vec   & 54.9 & 59.0 & 63.3 & 63.7 \\
 & MLRS       & 55.9 & 60.1 & \textbf{64.9} & \textbf{65.2} \\
 & \OurMethod & \textbf{59.6} & \textbf{64.2} & 62.5 & 63.1 \\
\bottomrule
\end{tabular}
\caption{Agreement ratio (Agr, \%) alongside accuracy (Acc, \%) on the two multiple-choice datasets (GMMLU and BELEBELE).
The two metrics induce the same ranking over methods in five of six model--dataset settings.}
\label{tab:agreement}
\end{table}

%% file: figures/sources/appendix/figure_steering_results_gmmlu.tex
\begin{figure}[t]
    \begin{minipage}{\columnwidth}
        \includegraphics[width=\columnwidth]{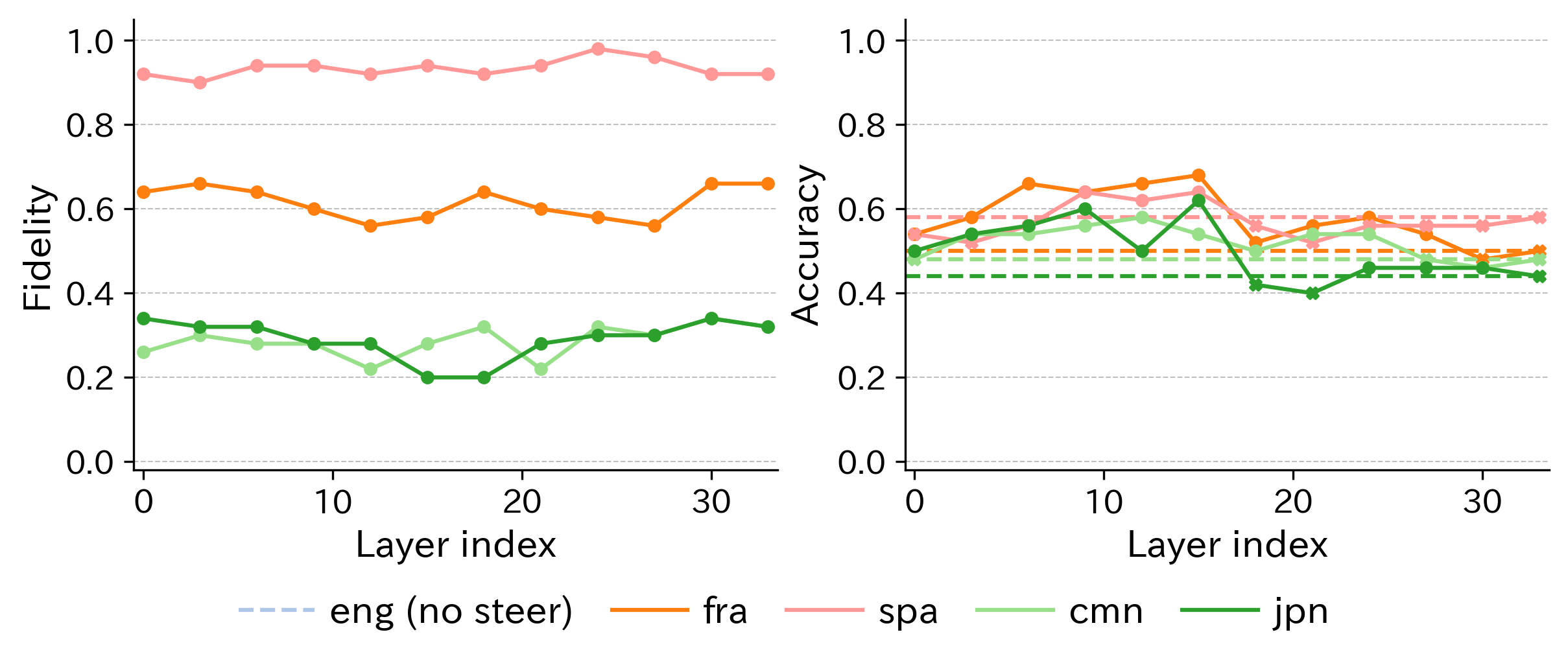}
        \subcaption{gemma3-4b}
    \end{minipage}
    \hspace{5mm}
    \begin{minipage}{\columnwidth}
        \includegraphics[width=\columnwidth]{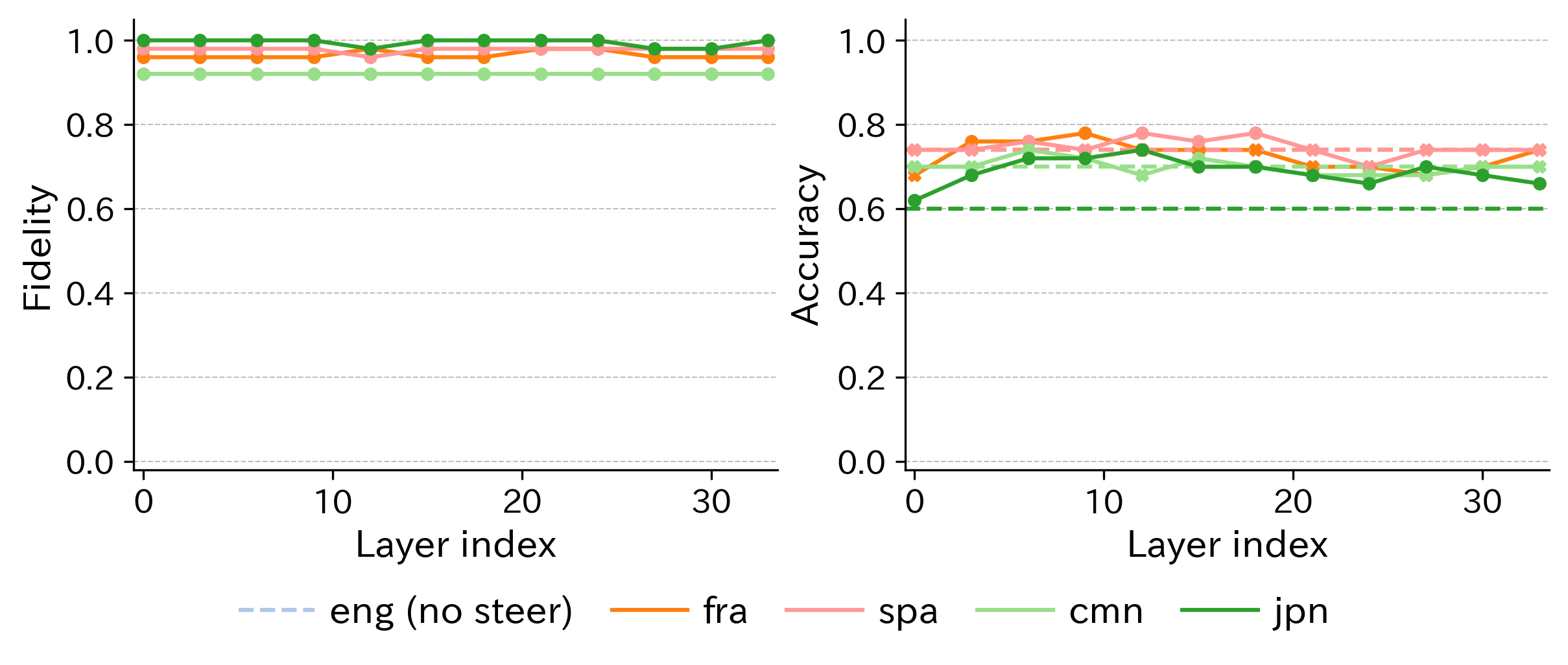}
        \subcaption{Qwen3-4B}
    \end{minipage}
  \caption{Fidelity (left) and accuracy (right) of gemma3-4b and Qwen3-4B on GMMLU dev when applying steering with \OurMethod.
The results of Llama3.1-8B are shown in \autoref{fig:llama_steering_results_gmmlu}.}
  \label{fig:gemma_qwen_steering_results_gmmlu}
\end{figure}

%% file: figures/sources/appendix/figure_langvec_steering_results_gmmlu_dev.tex
\begin{figure}[t]
    \begin{minipage}{\columnwidth}
        \includegraphics[width=\columnwidth]{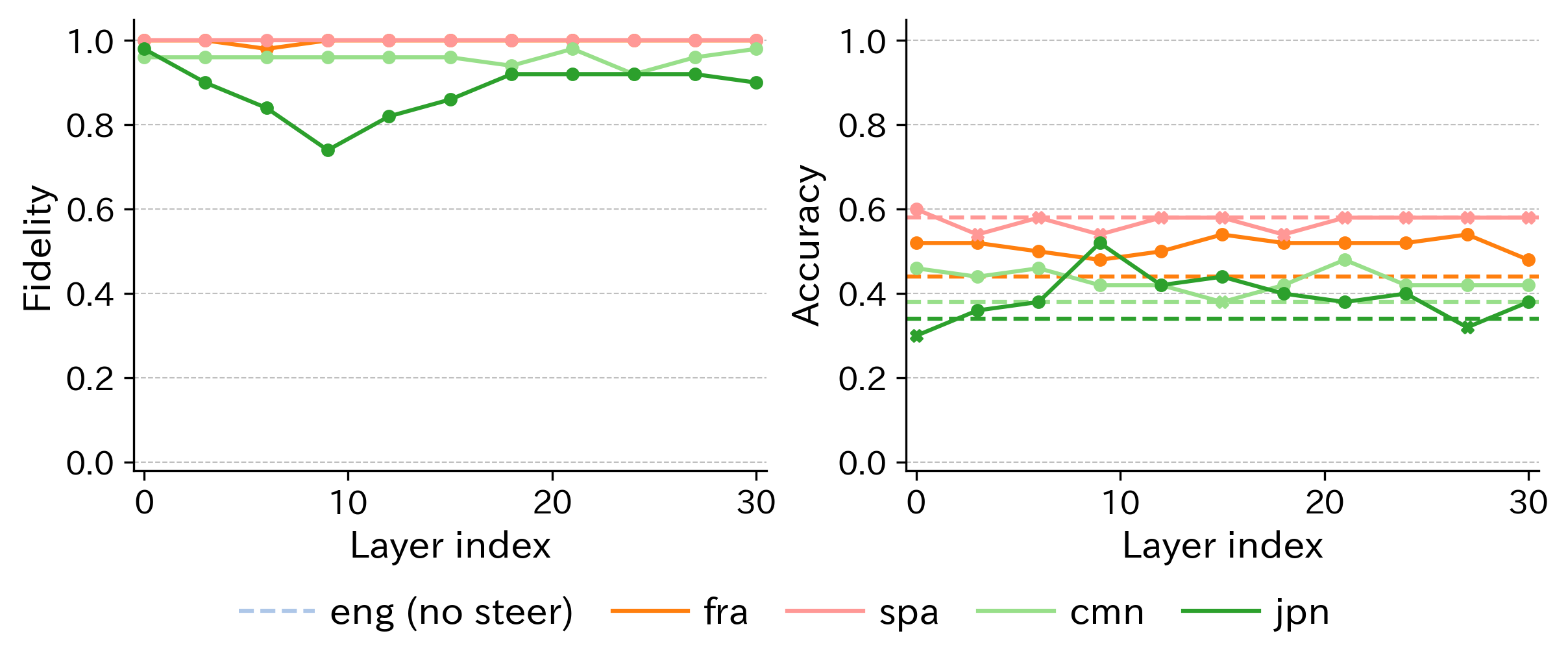}
        \subcaption{Llama3.1-8B}
    \end{minipage}
    \hspace{5mm}
    \begin{minipage}{\columnwidth}
        \includegraphics[width=\columnwidth]{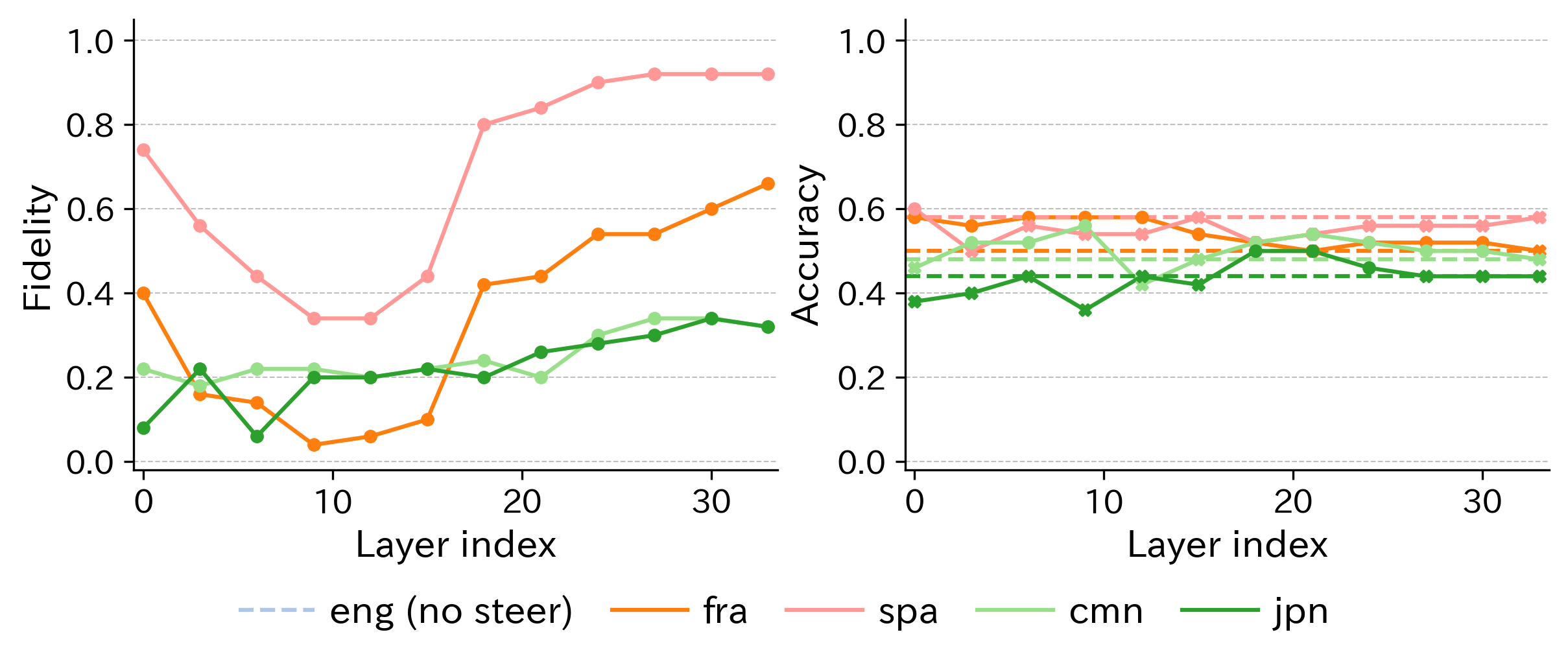}
        \subcaption{gemma3-4b}
    \end{minipage}
    \hspace{5mm}
    \begin{minipage}{\columnwidth}
        \includegraphics[width=\columnwidth]{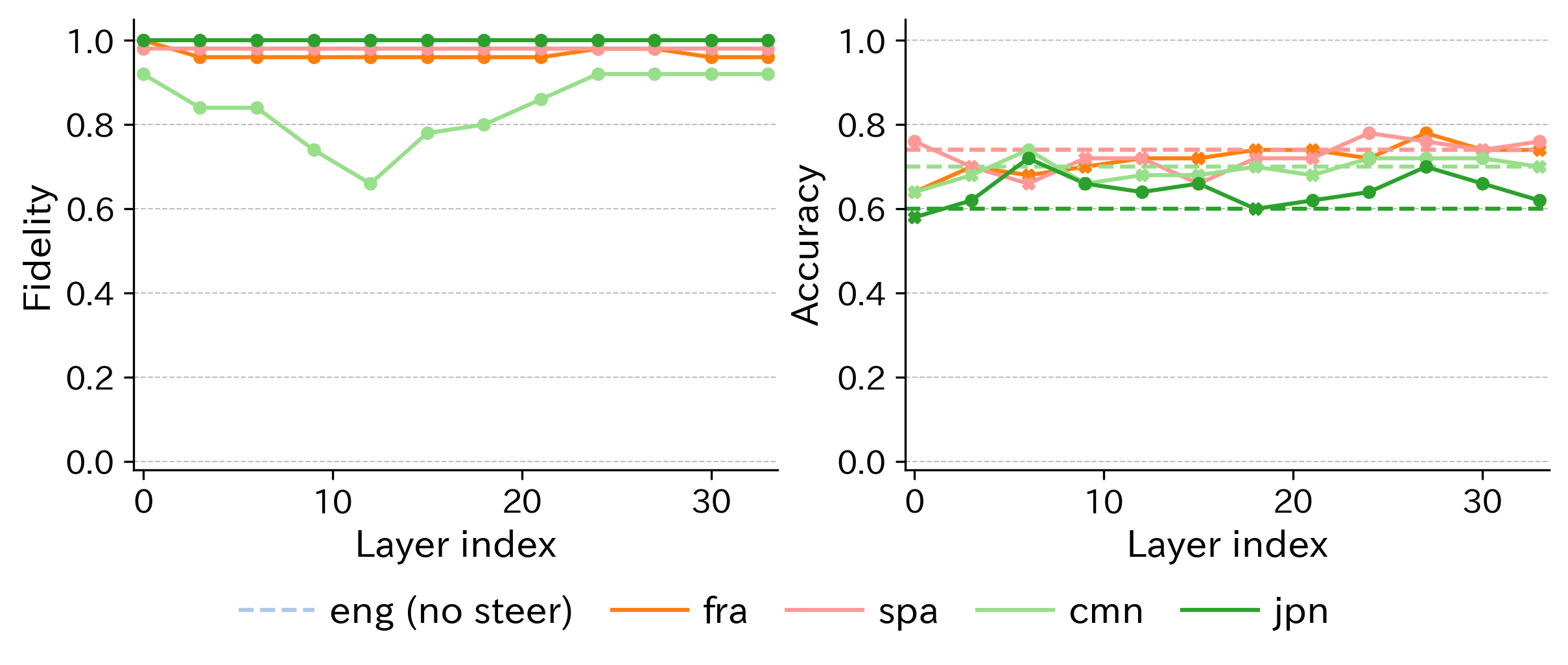}
        \subcaption{Qwen3-4B}
    \end{minipage}
  \caption{Fidelity (left) and accuracy (right) of Llama3.1, gemma3-4b and Qwen3-4B on GMMLU dev when applying steering with \textbf{Lang Vec}.}
  \label{fig:llama_gemma_qwen_steering_results_lang_vec_gmmlu_dev}
\end{figure}

%% file: figures/sources/appendix/table_ablation_steering.tex
\begin{table*}[t]
\centering
\small
\setlength{\tabcolsep}{6pt}

\begin{tabular}{@{}ll cc cc cc cc@{}}
\toprule
& & \multicolumn{2}{c}{GMMLU (9)} & \multicolumn{2}{c}{KLAR (7)} & \multicolumn{2}{c}{BELEBELE (9)} & \multicolumn{2}{c}{XQuAD (5)} \\
\cmidrule(lr){3-4} \cmidrule(lr){5-6} \cmidrule(lr){7-8} \cmidrule(lr){9-10}
Model & Method & Acc & Fid & Acc & Fid & Acc & Fid & Acc & Fid \\
\midrule
\multirow{4}{*}{\textit{Llama3.1-8B}}
 & \OurMethod       & 49.8 & 93.1 & 73.3 & 93.8 & 81.9 & 85.3 & \textbf{41.2} & 72.7 \\
 & No LEACE         & \textbf{55.7} & 91.3 & \textbf{73.6} & 92.5 & 81.4 & 82.7 & 39.6 & 69.9 \\
 & No Filter        & 55.5 & 91.7 & 71.9 & 91.6 & \textbf{82.1} & 85.5 & 34.7 & 71.3 \\
 & Last Token       & 49.5 & 93.5 & 67.8 & 93.6 & 76.5 & 88.9 & 36.2 & 78.0 \\
\midrule
\multirow{4}{*}{\textit{gemma3-4b}}
 & \OurMethod       & 51.8 & 62.6 & 75.8 & 96.4 & \textbf{66.0} & 92.9 & 26.0 & 82.5 \\
 & No LEACE         & 49.4 & 63.2 & 67.9 & 92.5 & 64.1 & 94.1 & 21.2 & 86.2 \\
 & No Filter        & \textbf{53.2} & 59.0 & \textbf{77.2} & 93.3 & 65.1 & 87.9 & \textbf{31.9} & 75.8 \\
 & Last Token       & 49.4 & 63.6 & 67.7 & 94.5 & 64.7 & 93.8 & 22.0 & 86.2 \\
\midrule
\multirow{4}{*}{\textit{Qwen3-4B}}
 & \OurMethod       & 59.6 & 98.1 & \textbf{68.0} & 90.4 & 62.5 & 89.5 & 45.6 & 69.9 \\
 & No LEACE         & 57.2 & 98.3 & 60.4 & 92.2 & 63.6 & 89.7 & 44.7 & 71.7 \\
 & No Filter        & \textbf{59.8} & 97.3 & 54.0 & 87.2 & 57.8 & 91.5 & 36.8 & 71.2 \\
 & Last Token       & 57.0 & 98.4 & 56.8 & 93.9 & \textbf{64.7} & 89.5 & \textbf{46.1} & 71.5 \\
\bottomrule
\end{tabular}
\caption{Ablation results of steering with \OurMethod, \OurMethod without LEACE-based decomposition (No LEACE), \OurMethod without perplexity-based filtering of the OT plan (No Filter), and swapping the language-agnostic representations of only the last token between the source and target languages (Last Token).
Bold marks the highest value per metric group.}
\label{tab:steering_ablation}
\end{table*}

%% file: figures/sources/appendix/figure_llama_training_cossim.tex
\begin{figure}
    \centering
    \includegraphics[width=\linewidth]{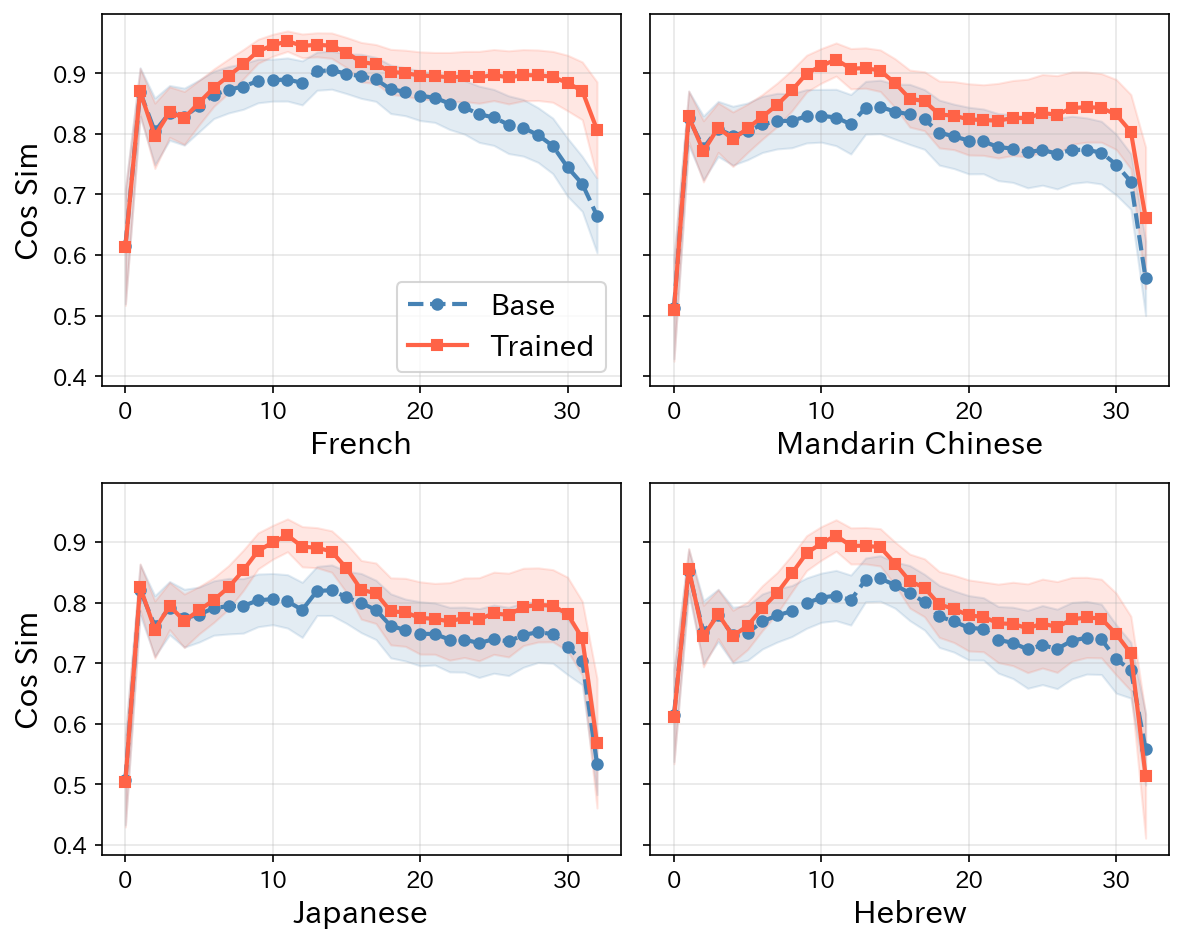}
    \caption{Cosine similarity between the source and target language-agnostic representations of Llama3.1-8B across layers before and after training with \OurMethod, calculated on FLORES+ \textit{devtest} set.}
    \label{fig:llama_training_cossim}
\end{figure}